\documentclass[preprints,article,accept,moreauthors,pdftex]{Definitions/mdpi} 
\firstpage{1} 
\pubvolume{1}
\issuenum{1}
\articlenumber{0}
\pubyear{2021}
\copyrightyear{2021}
\datereceived{} 
\dateaccepted{} 
\datepublished{} 
\hreflink{https://doi.org/} 

\makeatletter		
\let\c@lofdepth\relax		
\let\c@lotdepth\relax		
\makeatother
\usepackage{subfigure}
\makeatletter
\renewcommand{\@thesubfigure}{\normalsize(\textbf{\alph{subfigure}})}
\makeatother

\Title{RIANN---A Robust Neural Network Outperforms Attitude Estimation Filters}

\TitleCitation{RIANN---A Robust Neural Network Outperforms Attitude Estimation Filters}

\Author{Daniel Weber $^{1,}$* \orcidA{}, Clemens Gühmann $^{1}$ and Thomas Seel $^{2}$}

\AuthorNames{Daniel Weber, Clemens Gühmann and Thomas Seel}

\AuthorCitation{Weber, D.; Gühmann, C.; Seel, T.}

\address{%
$^{1}$ \quad Chair of Electronic Measurement and Diagnostic Technology, Technische Universität Berlin, Berlin 10587
 , Germany; d.weber.1@tu-berlin.de (D.W.); clemens.guehmann@tu-berlin.de (C.G.) \\
$^{2}$ \quad  Department Artificial Intelligence in Biomedical Engineering, Friedrich-Alexander-Universität Erlangen-Nürnberg, Erlangen 91052, Germany; thomas.seel@fau.de}

\corres{Correspondence: d.weber.1@tu-berlin.de}

\abstract{Inertial-sensor-based attitude estimation is a crucial technology in various applications, from human motion tracking to autonomous aerial and ground vehicles. Application scenarios differ in characteristics of the performed motion, presence of disturbances, and environmental conditions. Since state-of-the-art attitude estimators do not generalize well over these characteristics, their parameters must be tuned for the individual motion characteristics and circumstances. We propose \textit{RIANN}, a ready-to-use, neural network-based, parameter-free, real-time-capable inertial attitude estimator, which generalizes well across different motion dynamics, environments, and sampling rates, without the need for application-specific adaptations. We gather six publicly available datasets of which we exploit two datasets for the method development and the training, and we use four datasets for evaluation of the trained estimator in three different test scenarios with varying practical relevance. Results show that \textit{RIANN} outperforms state-of-the-art attitude estimation filters in the sense that it generalizes much better across a variety of motions and conditions in different applications, with different sensor hardware and different sampling frequencies. This is true even if the filters are tuned on each individual test dataset, whereas \textit{RIANN} was trained on completely separate data and has never seen any of these test datasets. \textit{RIANN} can be applied directly without adaptations or training and is therefore expected to enable plug-and-play solutions in numerous applications, especially when accuracy is crucial but no ground-truth data is available for tuning or when motion and disturbance characteristics are uncertain. We made \textit{RIANN} publicly available.}


\begin{document}

\section{Introduction}

As a result of rapid improvements in microelectromechanical systems technologies, miniature Inertial Measurement Units (IMUs) have become more and more lightweight and small at reasonable accuracies. They have thus entered a wide range of applications in which some form of motion tracking or analysis is required. Popular examples are found in aerospace engineering, autonomous vehicle technologies, robotics, and~wearables for health and sports applications
~\cite{seel_inertial_2020}. 

To estimate the motion of an object from the raw readings of an IMU, one needs to determine the orientation of the sensor frame with respect to the vertical axis and horizontal plane, i.e.,~the attitude. While the attitude itself is of high interest in many applications (see e.g.,~\cite{euston_complementary_2008,ding_tricycle_2018,valarezo_anazco_hand_2021,marco_nonlinear_2018}), attitude estimation is also a crucial step in velocity and position strapdown integration since it enables the separation of gravitational acceleration and the change of velocity~\cite{woodman_introduction_2007}. 

It should be noted that often additional value lies in estimating the heading with respect to the local magnetic field from magnetometer readings. However, abundant research shows that the assumption of a homogeneous magnetic field is often violated~\cite{nazarahari_40_2021,de_vries_magnetic_2009}, and~a wide range of magnetometer-free methods and solutions rely on deriving attitude estimates directly from gyroscope and accelerometer readings~\cite{kok_optimization-based_2014,teufl_validity_2018,lorenz_approach_2020,eckhoff_sparse_2020,grapentin_sparse_2020,lehmann_magnetometer-free_2020}. This is sometimes called 6D (or 6-axis) sensor fusion, in~contrast to 9D (or 9-axis) sensor fusion, which also incorporates the three-dimensional magnetometer readings. Any solution to the latter must contain a solution to the~former.

\subsection{The Challenge of Generalizability in Inertial Attitude~Estimation}

A multitude of advanced methods has been proposed for inertial attitude estimation via 6D sensor fusion. The~vast majority of these methods are complementary filters or Kalman filters with various modifications~\cite{nazarahari_40_2021}. All of these algorithms have parameters such as covariance matrices and filter gains that are used to adjust the filter to different measurement errors and motion characteristics. When an algorithm is applied to a given application problem, it is highly desirable to use it plug-and-play without any adaption of the mentioned parameters. However, given the wide variety of these conditions in different applications, state-of-the-art attitude estimators need to be tuned for every application to assure small errors~\cite{nazarahari_40_2021}. This represents a significant lack of generalization ability across different motion characteristics, environmental conditions, and~application demands. A~recent comparison of ten different state-of-the-art filters on a dataset with three different rotation speeds has shown that filters with more tunable parameters have the potential for lower errors on a specific task but perform worse without a suitable parameter selection~\cite{caruso_analysis_2021}. For~example, in~a fast and jerky motion, the~accelerometer must be used much more carefully than during a smooth and slow motion~\cite{laidig_broad_nodate}.

It was found that the optimal parameter regions for different motion characteristics rarely overlap, that default parameterizations often yield inadequate results, and~that specific tuning is critical for many experimental scenarios~\cite{caruso_analysis_2021}. To~date, there is no attitude estimator that performs robustly well (i.e., without requiring parameter tuning) across the different motion characteristics, sensor hardware, sampling rates, and~environmental conditions that appear in different application~scenarios.

\subsection{The Potential of Neural Networks in Inertial Attitude~Estimation}
The history of 70 years of AI research has shown that in many applications leveraging human understanding granted a short-term boost to performance and efficiency, but~in the long-term more general approaches that require more computing resources and data often succeeded, for~example in computer vision and natural language processing~\cite{rich_bitter_2019}. Inspired by this observation, an~alternative approach to the given attitude estimation task is to train a neural network end-to-end on the raw IMU data of a large variety of experimental datasets with ground truth measurements. Considering the success of neural networks in other system identification tasks~\cite{andersson_deep_2019,oord_wavenet_2016}, it seems promising to employ them for robust attitude~estimation.

To date, neural networks have been used to augment conventional attitude estimation methods by classifying the type of motion~\cite{brossard_rins-w_2020}, compensating measurements errors~\cite{brossard_denoising_2020}, or~smoothing the output of the conventional filter~\cite{chiang_artificial_2009,dhahbane_neural_2021,al-sharman_deep-learning-based_2020}. In~\cite{esfahani_aboldeepio_2019,esfahani_orinet_2020} Recurrent Neural Networks (RNNs) are used end-to-end for the strapdown-integration but still require additional sensor fusion to be applicable in long-term attitude estimation tasks. All of these methods improve the estimation performance by optimizing the estimator for a specific task in contrast to making it more robust across different tasks. A~first step in that direction has been taken by the authors of the current contribution in a recent conference paper~\cite{weber_neural_2020}. Therein, a~neural network structure with domain-specific adaptations has been developed and applied to a dataset with a wide variety of motions, demonstrating that neural networks can outperform conventional filters. However, the~study was limited to a single dataset with one specific sensor and one specific sampling rate, which strongly limits the usefulness of the neural network for attitude~estimation.

\subsection{Contributions}
The present work introduces \textit{RIANN} (Robust IMU-based Attitude Neural Network), a~ready-to-use, real-time capable, neural-network-based attitude estimator with no need for task- or condition-specific tuning. The~main contributions~are:
\begin{itemize}
    \item We propose three domain-specific advances for neural networks in the context of inertial attitude estimation.
    \item We identify two methods that enable neural networks to handle different sampling rates in system identification tasks.
    \item We present the attitude estimation neural network \textit{RIANN}, which results from these advances, and~make it publicly available at~\cite{weber_riann_2021}.
    \item We combine six different publicly available datasets for a comprehensive evaluation of the robustness of attitude estimation methods. 
    \item We compare \textit{RIANN} with commonly used state-of-the-art attitude estimation filters in three evaluation scenarios with different degrees of practical relevance. 
    \item We show that \textit{\textit{RIANN}} consistently outperforms commonly used state-of-the-art attitude estimation filters across different applications, motion characteristics, sampling rates, and~sensor hardware.    
\end{itemize}

\section{Problem~Statement}
\label{sec:problem}
The problem that is addressed by the present work is to design an attitude estimator that processes the gyroscope and accelerometer measurements of an IMU to provide real-time estimates of the sensor's attitude with respect to the vertical axis defined by Earth's gravitational field. In~the following, we give a precise definition of the problem and the performance metric by which any solution to that problem can be~assessed. 

\begin{figure}[H]
    \includegraphics[width=0.65\textwidth]{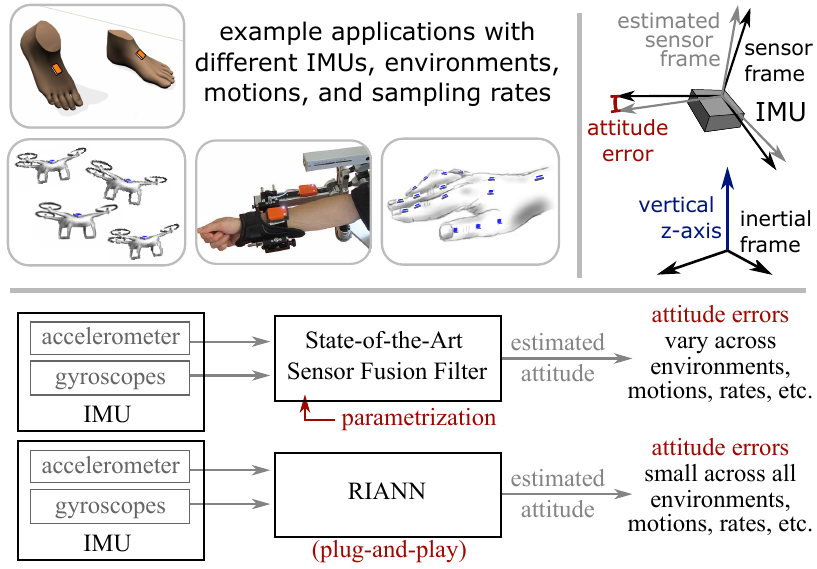}
    \caption{IMUs are used in various applications to measure an object's attitude with respect to the vertical axis. A~robust attitude estimator, unlike conventional filters, performs well across the different sensor hardware, motion characteristics, environmental conditions, and~sampling rates without application- or trial-specific parameter tuning. Graphic based on
    ~\cite{weber_neural_2020,beuchert_overcoming_2020} }
    \label{fig:visual_abstract}
\end{figure}

Consider the fundamental problem of attitude estimation, in~which an inertial sensor with a right-handed coordinate system $\mathcal{S}$ is rigidly attached to an object of interest. For~any motion that the object of interest performs, we strive to estimate the sensor's attitude, i.e.,~the orientation of the frame $\mathcal{S}$ with respect to the vertical axis. That estimation should be based on current and previous (but not future) measurement samples $\mathbf{a}(t_k)$ 
 and $\mathbf{g}(t_k)$ of the three-dimensional accelerometers and gyroscopes, respectively, i.e.,~we consider a filtering problem and omit magnetometer readings. See Figure~\ref{fig:visual_abstract} for~illustration.

Unlike many previous works, we refrain from assuming an initial rest period for filter convergence, since we deem this assumption too restrictive for a range of application scenarios. For~the same reason, we assume that the inertial sensor is factory-calibrated, but~no dedicated calibration of the turn-on bias has been performed. Such bias calibration algorithms typically also require static periods, which are difficult to assure and restrictive to assume in many applications. Instead, we consider the non-restrictive setting in which the estimation task is initialized during some motion with arbitrary rotation and translation characteristics, and~the available gyroscope and accelerometer measurements exhibit standard noise and bias~errors.

We formalize the given attitude estimation task using the mathematical notion of unit quaternions, which avoids the singularities in Euler angles. Let $\mathcal{E}$ be some inertial frame whose z-axis $\mathbf{e}_z=[0,0,1]^\intercal$ is aligned with the vertical axis, i.e.,~we neglect Earth's rotation. Represent the relative orientation between $\mathcal{S}$ and $\mathcal{E}$ as a unit quaternion $\mathbf{q}$ with the components $[w,x,y,z]$ and~assume that an estimate $\hat{\mathbf{q}}$ of that relative orientation $\mathbf{q}$ is provided by some attitude estimation algorithm. If~$\hat{\mathbf{q}}$ correctly describes the sensor's attitude, then $\hat{\mathbf{q}}$ equals ${\mathbf{q}}$ up to some \emph{heading} rotation around the vertical axis, which implies that the rotation axis of the error quaternion
\begin{align}
 \mathbf{q}_\text{err}:=& \mathbf{q}\otimes\left(\mathbf{\hat{q}}^{-1}\otimes\mathbf{q}\right)\otimes\mathbf{q}^{-1} \\
                       =& \mathbf{q}\otimes\mathbf{\hat{q}}^{-1}
\end{align}
is exactly the \emph{z}-axis. Note the important detail that $\mathbf{q}_\text{err}$ is defined and determined in \mbox{$\mathcal{E}$ coordinates.}

In the more general case of a non-zero attitude estimation error, a~scalar measure is needed that quantifies the disagreement between the true and the estimated attitude regardless of the aforementioned heading difference. To~this end, note that every error quaternion $\mathbf{q}_\text{err}$ can be decomposed into a heading error and an attitude error, i.e.,~into a rotation $\mathbf{q}_\text{head\,err}$ around the vertical axis and the smallest possible rotation $\mathbf{q}_\text{att\,err}$ around any horizontal axis. That smallest rotation angle can be determined analytically~\cite{laidig_broad_nodate} at any sampling instant $t_k$ and for any given $\mathbf{q}_\text{err}(t_k) =[w,x,y,z]$ by
\begin{align}
 e_\alpha(t_k) = 2\arccos\sqrt{w^2+z^2},
 \label{eq:e_a}
\end{align}
and it is equal to the angle between the true vertical axis ${\mathbf{q}}\otimes\mathbf{e}_z\otimes{\mathbf{q}}^{-1}$ and the estimated vertical axis $\hat{\mathbf{q}}\otimes\mathbf{e}_z\otimes\hat{\mathbf{q}}^{-1}$. We can therefore use $e_\alpha(t_k)$ to correctly quantify the disagreement between the true attitude and any estimated~attitude.

In the following, we consider established and novel methods that solve the given attitude estimation problem and quantify their performance by the root-mean-square of $e_\alpha(t_k)$ over the duration of motion in many different non-restrictive~scenarios.

\section{Neural Network Structure and~Implementation}
\label{sec:nn}
In this section, we present the current state-of-the-art methods for common time series regression that are suitable for application to the attitude estimation task. Based thereon, we propose domain-specific advances, which lead to a neural network that will be trained and studied in Section~\ref{sec:model}.

\subsection{Choice of the Neural Network~Structure}
When addressing the given problem by means of artificial neural networks, several different network structures might be considered. The~main candidates for processing time-series data are Temporal Convolutional Networks (TCNs), Transformers, and~Recurrent Neural Networks (RNNs).

TCNs are stateless feed-forward neural networks~\cite{andersson_deep_2019}, which are able to model dynamic systems by processing windows of a fixed size at once instead of samples sequentially. Transformers are the current state-of-the-art architectures for natural language processing, because~of their ability to process relations between two distant points in time~\cite{wolf_huggingfaces_2020}.

RNNs have recurrent connections in their hidden layers, which store state information between time steps. The~main advantage of this approach is that the calculation is very efficient and the state information may be stored infinitely in theory. In~practice, there are limits to the number of time steps that may be performed before the state has degraded too much, because~of the vanishing gradient problem~\cite{gonzalez_non-linear_2018}. Targeting this issue, many RNN architectures have been developed with Long Short-Term Memories and more recently Gated Recurrent Units (GRUs) \cite{cho_properties_2014} being the most common one. They use a gating mechanism to alleviate the numerical problems, allowing for training with thousands instead of hundreds of time-steps in one mini-batch. The~inherently sequential nature of RNNs limits the parallelizability of the training and especially the~inference.

Previous work has shown that the RNN variant GRU outperforms TCNs in the attitude estimation task because of its ability to store state information over an indefinite amount of time~\cite{weber_neural_2020}. Transformers on the other hand have similar capabilities but are less suited to real-time applications in environments with limited resources because of their large amount of required memory and computing capacity. Therefore, we use GRUs to process the sequential~signal. 

\begin{figure}[H]
    \subfigure[~neural network structure]{\includegraphics[width=0.74\textwidth]{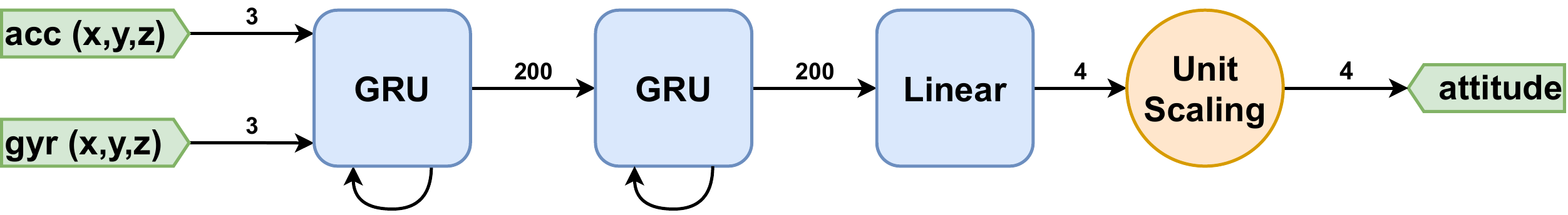}}
    \subfigure[~grouped-input neural network structure]{\includegraphics[width=0.74\textwidth]{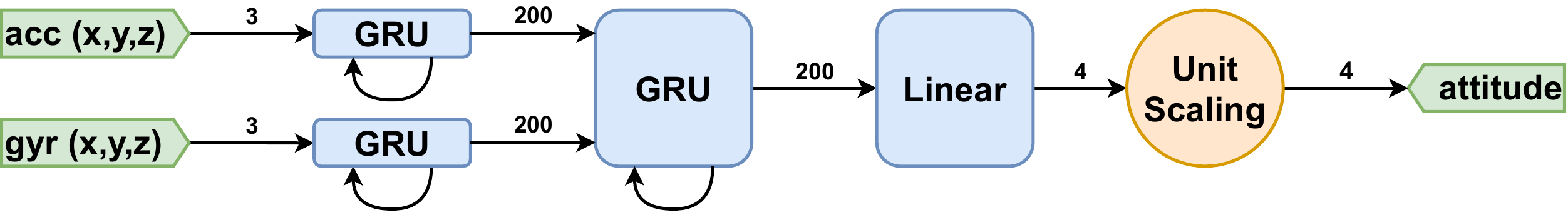}}
    \subfigure[~time-aware neural network structure]{\includegraphics[width=0.74\textwidth]{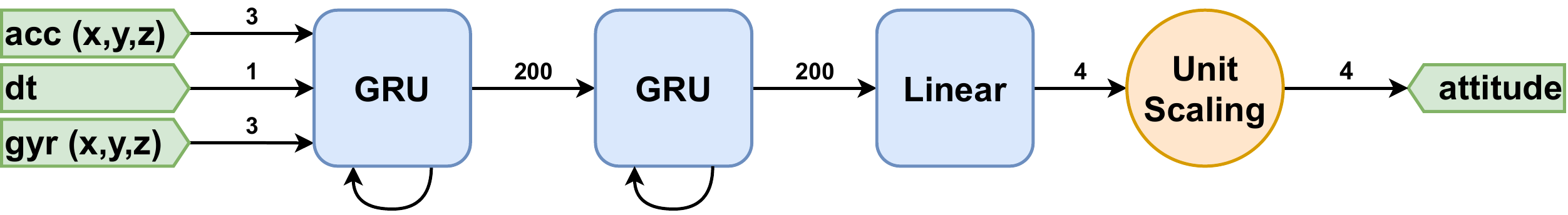}}
    \caption{The structure of the neural network (\textbf{a}); the grouped-input neural network (\textbf{b}) with separate layers for accelerometer and gyroscope; and the time-aware neural network (\textbf{c}) with the time-difference as an additional input for attitude~estimation.}
    \label{fig:gae_models}
\end{figure}

A stack of two GRU layers, which transforms the 6-dimensional IMU input $\mathbf{u}(k)$ of every sampling instant $t_k$ to an $N_n$-dimensional feature vector $\mathbf{h}(t_k)$, with~$N_n$ being the number of neurons per layer, has proven to be effective in attitude estimation~\cite{weber_neural_2020}. To~assure that the network output is a unit quaternion, the~$N_n$-dimensional feature vector $\mathbf{h}(t_k)$ is transformed to a four-dimensional vector $\mathbf{\hat{q}}(t_k)$ with a Euclidean norm of 1. To~this end, we use a linear layer with a weight matrix $\mathbf{W}$ for dimensional reduction and normalize \mbox{the result:}
\begin{align}
 \mathbf{\hat{q}}(t_k) = \frac{\mathbf{W} \cdot \mathbf{h}(t_k)}{|| \mathbf{W} \cdot \mathbf{h}(t_k) ||} 
\end{align}
The complete model structure is visualized in Figure~\ref{fig:gae_models}a.
 
\subsection{Neural Network Implementation with General Best~Practices}

We train and evaluate the neural networks with datasets that consist of multiple measured sequences of sensor and ground truth data. To~avoid memorizing the same sequences, long overlapping windows get extracted from the measured sequences, so the neural network has to start at different points in time. Because~of the vanishing gradient problem, RNNs can only be trained with a limited number of time steps per mini-batch. To~process longer sequences in training, truncated backpropagation through time is used~\cite{tallec_unbiasing_2017}. It is a method that splits sequences into a chain of shorter sequences, which are used sequentially for training with the network keeping its last hidden state between each mini-batch. To~improve training stability and remove any scaling-related input signal bias, the~signals are standardized to zero mean and a standard deviation of one~\cite{ioffe_batch_2015}.  
A crucial component of the training process is the optimizer. We use the current state-of-the-art combination of RAdam and Lookahead, which has proven effective in various tasks~\cite{liu_variance_2019,zhang_lookahead_2019}. The~implementation of all adaptations in the training process has been done with the Fastai 2 API, which is based on PyTorch~\cite{howard_fastai_2020}. Parameterization of the learning rate is critical for the optimization process. We use the learning rate finder heuristic proposed in~\cite{smith_cyclical_2017} for the maximum learning rate and cosine annealing for faster convergence~\cite{loshchilov_sgdr_2017}.

Neural Networks have many hyperparameters that span a vast optimization space. There are two state-of-the-art hyperparameter optimization algorithms: Population Based Training (PBT) \cite{jaderberg_population_2017} and Asynchronous Successive Halving Algorithm (ASHA) \cite{li_system_2020}. PBT is an evolutionary algorithm that trains a population of neural networks in parallel, relying on the survival of the fittest principle. ASHA on the other hand is an early stopping algorithm that utilizes the observation that most of the models that perform well at the final epoch also perform well early in the training process. This way the number of configurations that may be tested is increased by orders of magnitude. It has been shown that PBT performs better in reinforcement learning because it is able to learn a schedule of hyperparameters but~performs worse in supervised learning~\cite{li_system_2020}. ASHA has the main advantage that it is easy to use and stable. Therefore, we optimize the neural networks in this work with~ASHA.

\subsection{Loss~Function}
For the error component of the loss function that is minimized during the training process, we use the metric $e_\alpha(t_k)$ as defined in \eqref{eq:e_a}. Taking the mean square results in the loss function $e_\text{MSE}$ for a sequence of $N$ samples starting at some sampling instant $t_k$:
\begin{align}
    e_\text{MSE} &= \frac{1}{N}\sum_{\kappa=k}^{k+N-1}e_\alpha(t_\kappa)^2
\end{align}
As pointed out in previous work~\cite{weber_neural_2020}, the~gradient of the loss function grows unbounded as the optimization approaches the target argument $1$ of the $\arccos$ function, which results in numerical issues:
\begin{align} 
    \lim\limits_{a \rightarrow 1}{\frac{\mathrm{d}\arccos(a)}{\mathrm{d}a}}&= \lim\limits_{a \rightarrow 1} \frac{-1}{\sqrt{1-a^2}} = -\infty.
\end{align}
The solution approach~\cite{weber_neural_2020} was to replace $\arccos$ in the loss function by a linear term $1-a$ that keeps the monotonicity and correlation with the attitude. This avoids all numerical problems but leads to a discrepancy between loss function and evaluation metric~\cite{weber_neural_2020}. We, therefore, propose to tackle the numerical problems directly by increasing the floating-point precision for the calculation of $e_\alpha(t_k)$ to 64 bit and cutting values that are too close to 1. This results in a numerically stable and direct projection of the metric to the loss function at a negligible computational~cost.

Gaps in the ground truth time series are problematic for the training process of neural networks since continuous data is needed for gradient calculation. Such gaps, however, are commonly present in motion tracking datasets due to temporary occlusion of optical markers or other disturbances of the optical reference system. Since filling the gaps compromises the integrity of the ground truth, we mask out the corresponding time intervals when generating the~mini-batches.

\subsection{Generalization Across Sampling~Rates}
\label{sec:model_time}
In this work, the~neural network is targeted to work equally well in a broad range of scenarios with different sampling rates. To~allow a neural network to operate as a filter with varying sampling rates, we propose a just-in-time-resampling (JITR) network and a time-aware (TA) neural network which will be evaluated in Section~\ref{sec:resStudy}.

The JITR network incorporates the idea to adapt an existing neural network that has been trained with a fixed sampling rate to the application of a broad range of sampling rates. This is achieved by resampling the input signal to the sampling rate of the neural network and doing the same in reverse with its output. This approach has the advantage of being applicable to every existing neural network. On~the other hand, for~every inference step, two resampling steps are required, which increases the required computation time and latency. In~addition to that, more inference time steps have to be taken if the neural network has a higher sampling rate than the source signal, which increases the required computation time even more---or information is lost if the neural network has a lower sampling rate than the source~signal.

The time-aware neural network incorporates sampling rate-related information to its input, allowing it to be applied to signals of different sampling rates directly. The~time difference between two samples $dt$ is used as an additional input, as~visualized in \mbox{Figure~\ref{fig:gae_models}b}. Since $dt$ is provided for every time step, the~network is generally able to process signals with unevenly sampled data, but~we leave the analysis of this case for future work. The~time-aware neural network needs to be trained on data with a range of sampling rates that are expected to be used in inference time. Since neural networks are known to carry the risk of bad extrapolation beyond the range of training data, the~performance of the time-aware neural network is expected to degrade outside the range of sampling rates used for~training.

In both models, the~input and output data have to be resampled either in the training or in the inference process. The~measured acceleration and angular velocity may be resampled independently with a conventional discrete-Fourier-transformation-based method~\cite{virtanen_scipy_2020}. The~output and reference signals are unit quaternions, which means that processing components independently generally leads to leaving the feasible set. For~resampling quaternions, we thus use spherical linear interpolation~\cite{shoemake_animating_1985}.

\subsection{Data~Augmentation}
With data augmentation, the~size of the training data can be increased by using domain-specific information. With~this method the generalizability of a network trained with a limited dataset may be improved, which has been demonstrated in computer vision~\cite{perez_effectiveness_2017} and audio processing~\cite{xiaodong_cui_data_2015}. We propose two data augmentation transformations for the attitude estimation task: the virtual IMU rotation and the induction of artificial inertial measurement~errors.

For the virtual IMU rotation, we transform all accelerometer data, gyroscope data, and~the ground truth attitude data of a given time interval by rotating them with a fixed randomly generated unit quaternion. If~the original data was generated by moving an object with a mounted IMU, then this virtual rotation simulates the effect of attaching the IMU to the moving object in a different orientation. By~this data augmentation, the~network's inference capabilities become independent of the sensor-to-object orientation, which crucially enriches any training~dataset.

There are multiple kinds of errors in inertial measurement data that influence the accuracy of the attitude estimation task~\cite{zhang_impact_2020}. We model the two most notable: the measurement noise and the gyroscope bias. For~noise augmentation, we apply normally distributed noise with randomly generated standard deviations to each raw data sequence. The~standard deviations are generated separately for the accelerometer and the gyroscope for every sequence. This also introduces varying levels of reliability of the accelerometer and the gyroscope into the training data. For~the bias augmentation, an~individual, randomly generated but constant offset is applied to every axis of the gyroscope measurement. The~error augmentation methods add new hyperparameters to the training process, which may be picked either based on available measurement data or via a hyperparameter optimization with representative validation data, which is what we will do in Section~\ref{sec:model}.

\subsection{Grouped Input~Channels}
As an alternative method to putting all measured signals in the same first layer, we consider creating groups of signals that are processed in separate layers, which are then merged in the following one. This reduces the possible interactions between the signals, which may assist the neural network in focusing on the relevant relations between the signals. In~the attitude estimation task, the~first layer is split into an accelerometer and a gyroscope layer, such that the accelerometer layer may provide attitude information in slow movements and the gyroscope layer may focus on the strapdown integration during rapid movements over time. Related work employed such approaches but without analyzing the influence on the models' performance~\cite{zheng_time_2014,esfahani_aboldeepio_2019}, which is what we will do in Section~\ref{sec:ablStudy}.

\section{Neural Network~Optimization}
\label{sec:model}


In this section, we train the proposed recurrent neural network and compare different combinations of the domain-specific advances developed in Section~\ref{sec:nn} to find the best performing network configuration and~hyperparameters.

\begin{figure}[H]
    \includegraphics[width=0.45\textwidth]{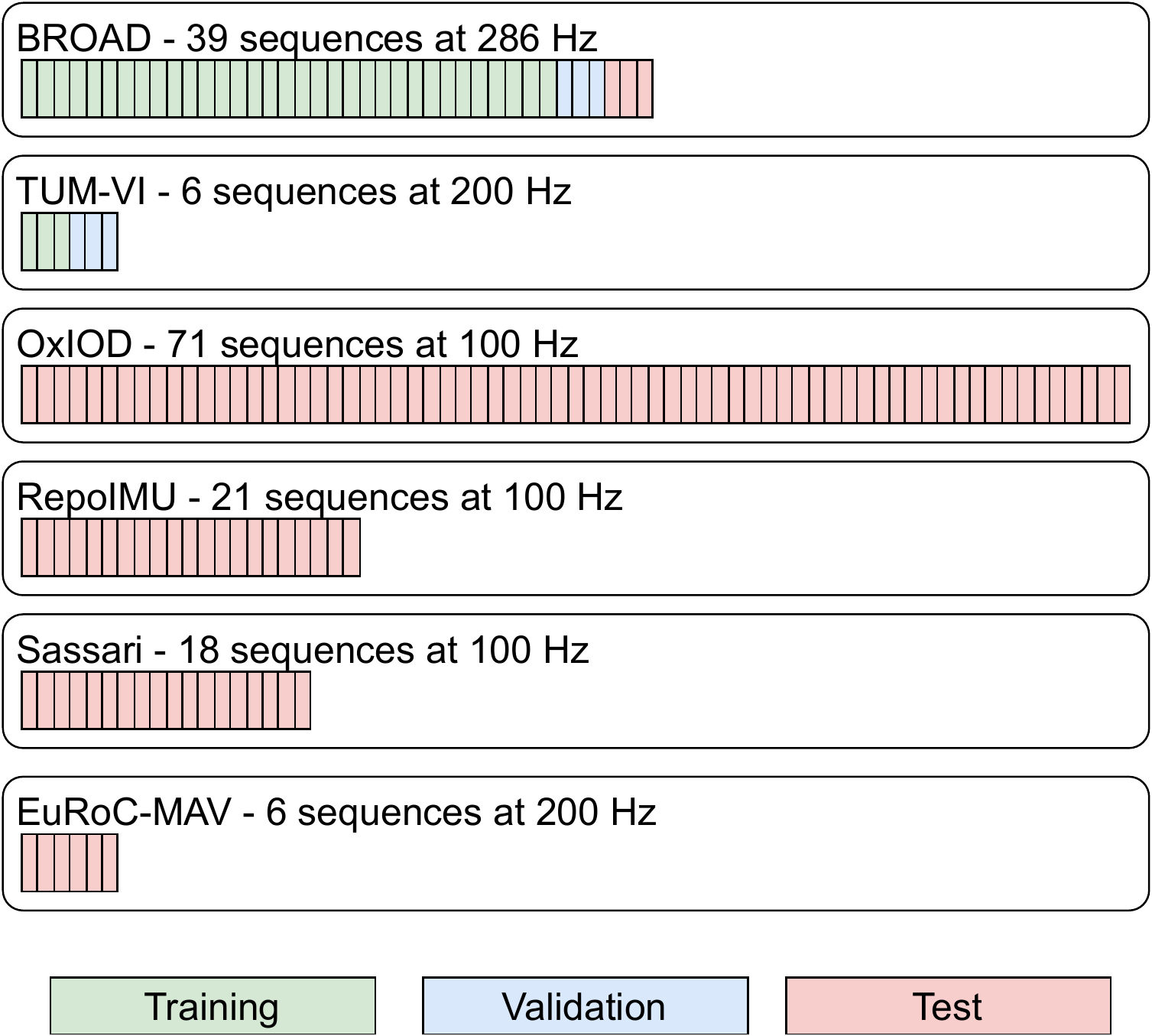}
    \caption{The dataset collection is composed of six publicly available datasets, which are split into training, validation, and~test data. While the validation data is used to find the best performing network configuration and hyperparameters, the~test data is reserved for the final performance evaluation in Section~\ref{sec:genAnalysis}.}
    \label{fig:datasetconfig}
\end{figure}

For the development of a robust network, we need a dataset with a wide spectrum of different motion characteristics. 
The dataset also needs to be large enough, so it can be split into training and validation data, which are used to optimize the hyperparameters and~test data, which is used for performance evaluation. We meet these requirements by combining six publicly available datasets with optical ground truths from different sources and application domains. Figure~\ref{fig:datasetconfig} shows the split of the combined dataset into training, validation, and~test data. The~BROAD dataset is an inertial dataset with a wide variety of motion characteristics~\cite{laidig_broad_nodate}. The~TUM-VI dataset contains inertial and optical measurements of a handheld camera rig moving in various environments, of~which we use the six room sequences because only they have an optical ground truth for the orientation over the whole sequence~\cite{schubert_tum_2018}. The~EuRoC-MAV dataset is composed of inertial and optical measurements on a micro aerial vehicle~\cite{burri_euroc_2016}. The~Sassari dataset is a rich inertial dataset with measurements of several different IMUs~\cite{caruso_mimu_optical_sassari_dataset_2020}. The~OxIOD Dataset is an inertial dataset with multiple devices and various types of motion~\cite{chen_oxiod_2018}. Finally, the~RepoIMU dataset comprises inertial measurements from motions of a T-stick and a pendulum~\cite{szczesna_reference_2016}.

\begin{figure}[H]
\vspace{-6pt}
    \centering
    \widefigure
    \includegraphics[width=\textwidth]{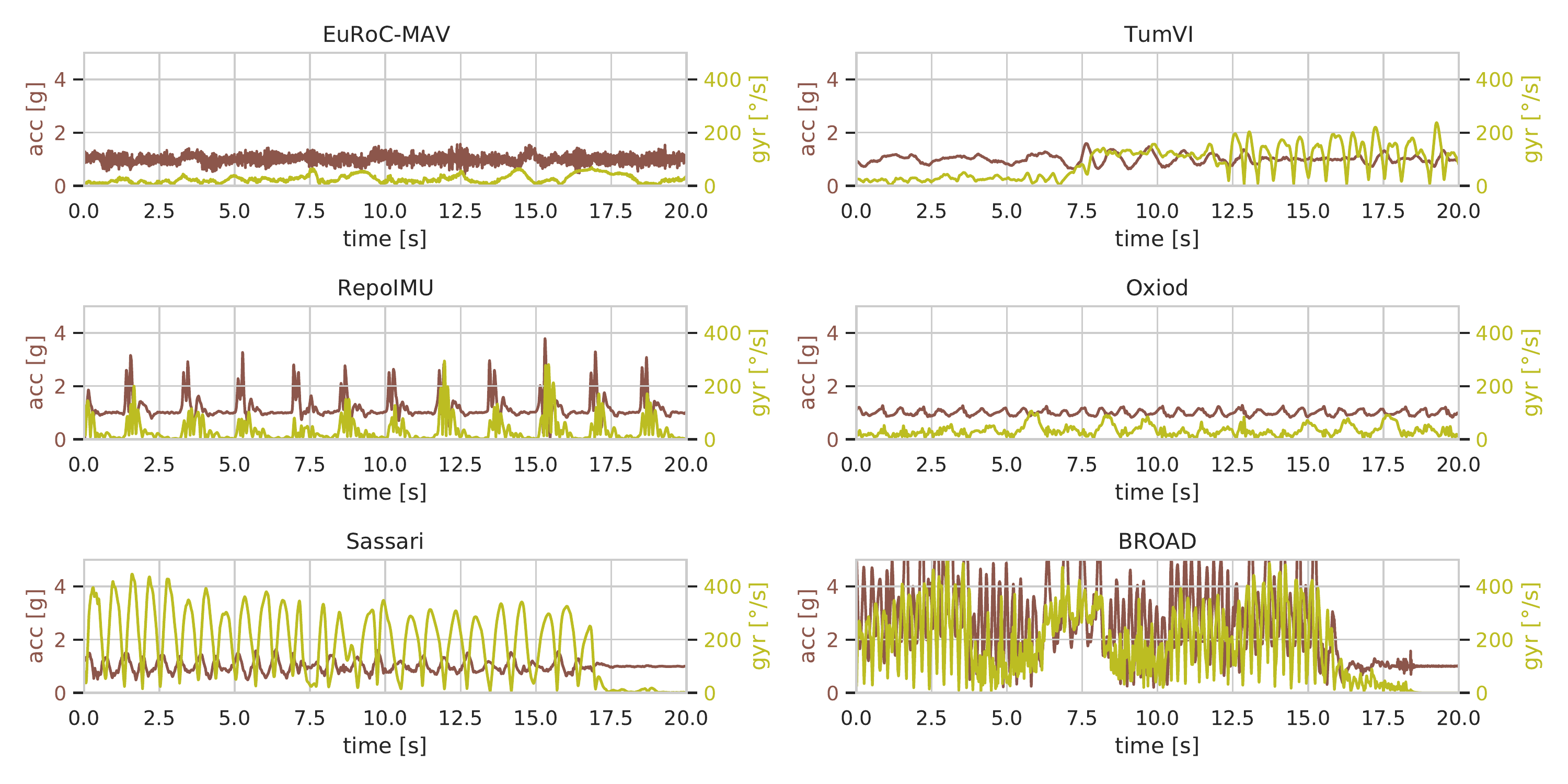}
    \caption{Raw signal magnitudes over time for exemplary sequences from all six datasets. Rotational and translational characteristics vary~largely.}
    \label{fig:gae_0_sequences}
\end{figure}
\begin{figure}[H]
    \centering
    \widefigure
    \subfigure[~standard deviation of measurements]{\includegraphics[width=0.48\textwidth]{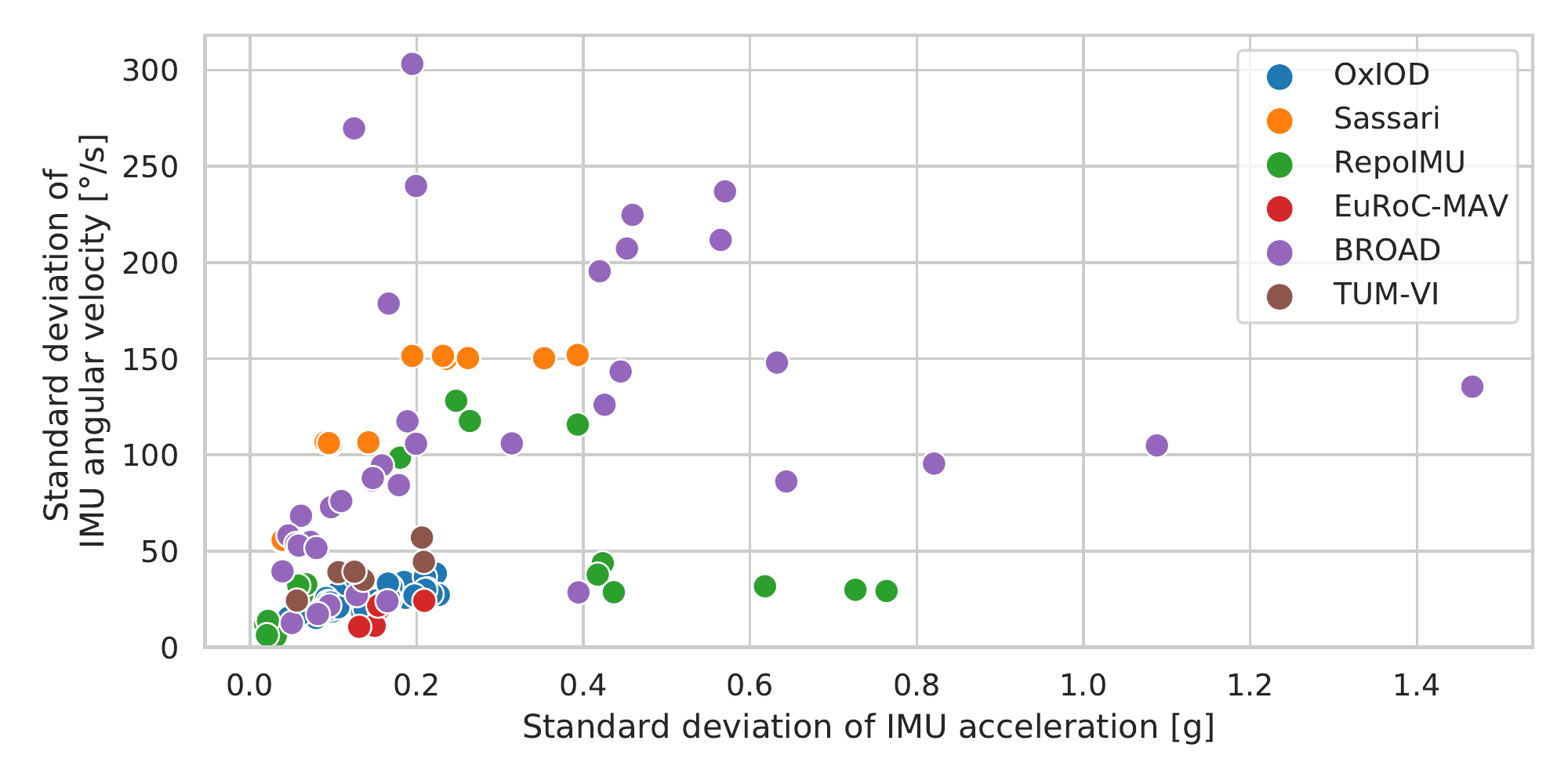}}
    \subfigure[~mean of measurements]{\includegraphics[width=0.48\textwidth]{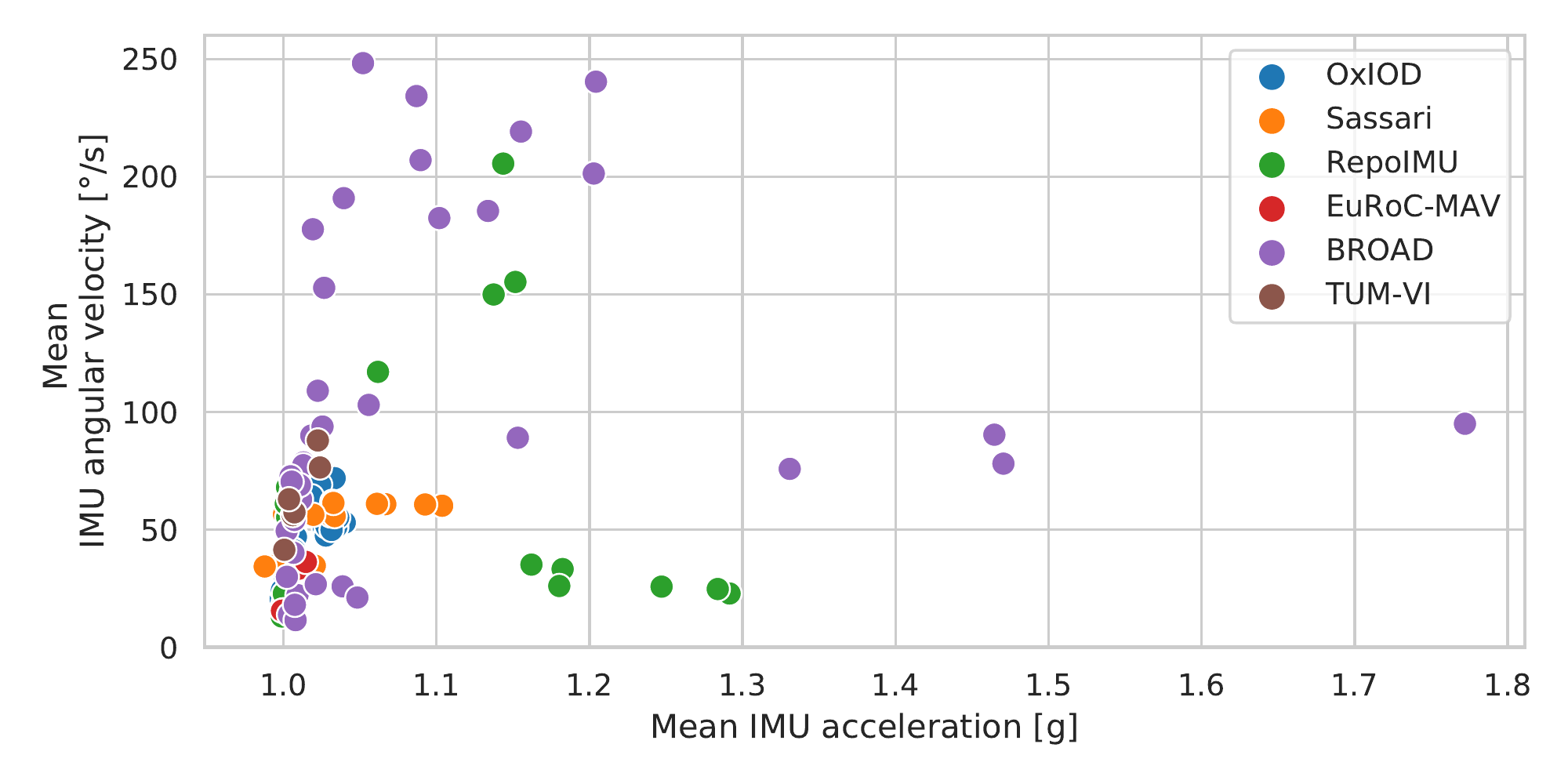}}
    \caption{Comparison of motion characteristics of the different datasets. Differences between datasets are considerable. The~datasets BROAD, Sassari, and~RepoIMU contain motions with a relatively large variety of~characteristics.}
    \label{fig:gae_0_scatter_plot}
\end{figure}

The datasets come from different applications with different motion patterns, on~which a \emph{robust} estimator should be able to work equally well without individual parameter tuning. Figure~\ref{fig:gae_0_sequences} illustrates the variety of motion characteristics in terms of one short exemplary time sequence from each dataset. The~entire spectrum of motion characteristics of all sequences of all datasets is visualized in Figure~\ref{fig:gae_0_scatter_plot} in terms of the mean and standard deviation of the accelerometer and gyroscope measurements. The~datapoints of most datasets create narrow clusters in dataset-specific regions of the plot, which demonstrates that most applications have a specific but limited spectrum of motion characteristics. This indicates that a sufficiently rich combination of data is required for the training of a robust neural network and, likewise, for~an evaluation that shows whether the network performs well across a broad range of scenarios. To~preserve as many datasets as possible for the evaluation of the final network in Section~\ref{sec:genAnalysis}, we decide to use only the BROAD dataset and the TUM-VI dataset for training and hyperparameter optimization in this~section.

The best network configuration is determined in three steps: ablation study, sampling rate study, and~network size analysis. While the ablation study quantifies the benefits of each domain-specific advance developed in Section~\ref{sec:nn}, the~sampling rate study identifies the best strategy for enabling the network to process data with a wide range of sampling rates. In~the network size analysis, we then quantify the effect of the parameter count on the estimation accuracy and~latency.

\end{paracol}
\nointerlineskip
\begin{paracol}{2}
\switchcolumn

\clearpage
\end{paracol}
\nointerlineskip
\begin{paracol}{2}
\switchcolumn

\vspace{-9pt}
\subsection{Ablation~Study}\label{sec:ablStudy}
To determine the best performing network configuration, we consider all combinations of a network with/without the loss function adaptation, with/without rotation augmentation, with/without error augmentation, and~with/without grouped input adaptation. This results in 16 possible network configurations. Each network configuration is trained on the training data and then applied to the validation data (cf. Figure~\ref{fig:datasetconfig}) to determine the average RMSE over all validation sequences. This process is repeated five times, and~the median of the five average RMSE values is used for comparison. To~exclude the sampling rate question from the described procedure, all training and validation time sequences are resampled to a fixed sampling rate of 300 Hz, which is chosen higher than all source sampling rates to avoid information loss in the resampling process. We include every time sequence once without and once with an artificial turn-on gyroscope bias, which was drawn from a normal distribution with a standard deviation of 0.5$\,^\circ/\text{s}$.

The results of the described comparison show that most of the proposed advances are sequentially dependent on each other and that successive improvements can be achieved as visualized in Figure~\ref{fig:gae_1_ablation}. A~naive state-of-the-art neural network for time series processing does not achieve competitive performance when compared to conventional attitude estimation filters. Optimizing the loss function to the task-specific requirements improves the results, but~the network does not generalize across different sensor-to-object orientations. The~proposed data augmentation by virtual rotations solves this problem and further improves the network performance. Adding also the error augmentation further decreases the error, whereas grouping the input brings no additional~benefit.

\begin{figure}[H]
\vspace{-6pt}
    \includegraphics[width=0.75\textwidth]{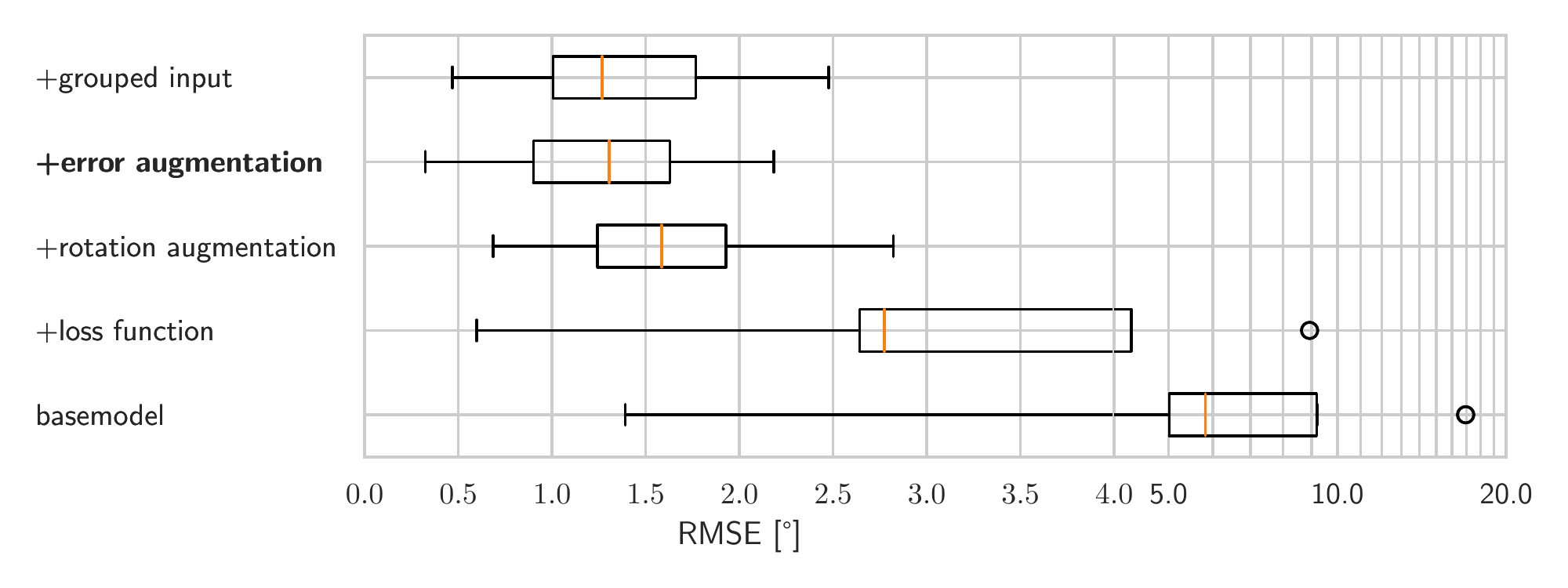}
    \caption{Ablation study of domain-specific advances, which are added successively (bottom to top). The~RMSE distributions over 12 validation sequences are compared. Adaptations of the loss function and data augmentation have the largest impact on accuracy. Bold font indicates the chosen final~configuration.}
    \label{fig:gae_1_ablation}
\end{figure}

All in all, the~best configuration is a state-of-the-art recurrent neural network for time series processing with an optimized loss function and data augmentation by virtual rotations and artificially induced measurement~errors.

\subsection{Sampling Rates~Study}\label{sec:resStudy}
The network configuration that was identified in the previous section performs well at a single sampling rate. We now combine that network with any of the two approaches that were proposed in Section~\ref{sec:model_time} for generalization to a broad range of sampling rates. More precisely, we first identify the best resampling strategy for the \emph{time-aware neural network} and then compare it to the \emph{JITR network}. The~study utilizes the training and validation data (cf. Figure~\ref{fig:datasetconfig}) resampled over a frequency range of 50 to 500 Hz, as~detailed below. To~compare different configurations, every configuration is trained five times, and~as before, the~median of the five average RMSE values is used for~comparison. 

For training the \emph{time-aware neural network}, each training sequence is resampled to a certain number $N_\text{sr}$ of different frequencies from the given range, which effectively multiplies the number of training sequences by $N_\text{sr}$. To~analyze the ability of the network to interpolate between sampling rate gaps, we consider training at $N_\text{sr}= 6$, 20, 100, or~500 different sampling rates. Additionally, we consider three different strategies for drawing these different sampling rates: equidistantly over the sampling time ($t_s$) space (2--20 ms), equidistantly over the sampling rate ($f_s$) space (50--500 Hz), or~both strategies combined. The~performance of these different configurations is compared in terms of the average RMSE over all validation sequences resampled to any frequency between 50 and 500 Hz, as~shown in Figure~\ref{fig:gae_2_sampling_fs}. In~the given frequency range, 20 different sampling rates or less lead to sub-optimal results for all resampling strategies. With~at least 100 different sampling rates, the~resampling with equidistant sampling rate values yields the lowest error over the entire frequency range. Since it achieved the lowest error, we denote the \emph{time-aware neural network} that was trained with 100 different equidistant sampling rates by \textit{NN-TA} and disregard the other resampling configurations in the~following.

\begin{figure}[H]
\vspace{-6pt}
    \includegraphics[width=0.75\textwidth]{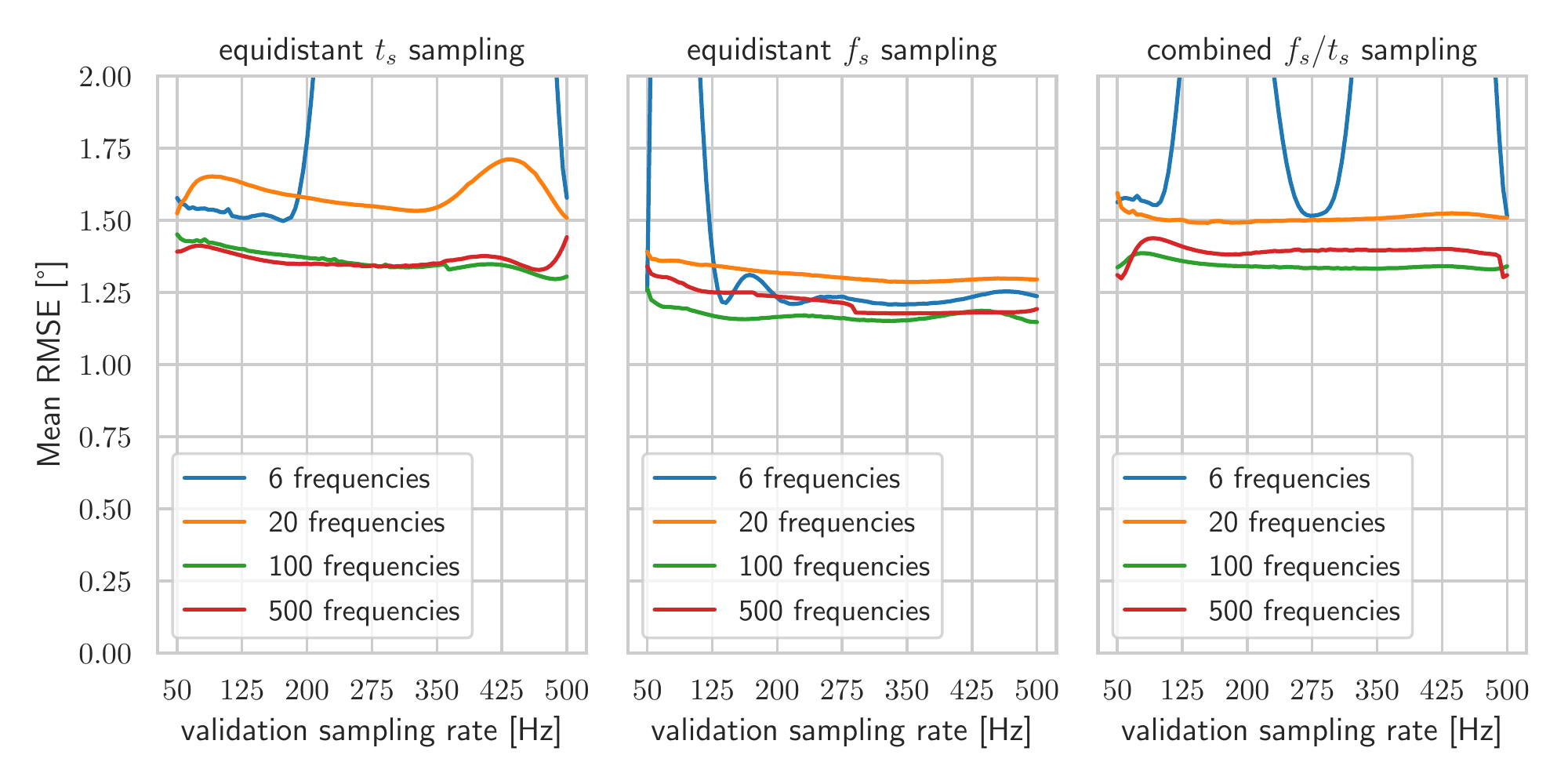}
    \caption{Comparison of time-aware networks with training data that was resampled to frequencies drawn equidistantly from sampling time $t_s$, sampling rate $f_s$, or~both combined. At~least 100 different frequencies are required to achieve high accuracy. The~$f_s$ space models are the most accurate across the entire frequency~range.}
    \label{fig:gae_2_sampling_fs}
\end{figure}

In the second step, we compare the network \textit{NN-TA} with the just-in-time resampling (JITR) approach from Section~\ref{sec:model_time}. We consider the neural network that resulted from the ablation study, add a JITR of the network's input data, and~denote that combination by \textit{NN-JITR}. Figure~\ref{fig:gae_2_sampling_comp} visualizes the mean RMSE of \textit{NN-TA} and \textit{NN-JITR} over a frequency range of 30 to 600 Hz, which is broader than the range of 50 to 500 Hz on which \textit{NN-TA} has been trained. \textit{NN-JITR} has a stable accuracy over the complete frequency range, whereas the performance of \textit{NN-TA} degrades outside of its training frequency range. However, inside that training range, \textit{NN-TA} performs better than \textit{NN-JITR}, which is probably due to the regularization introduced by the resampling in the training process. Considering the inference time benefits of the time-awareness approach, \textit{NN-TA} seems better suited for applications in embedded systems with sampling rates within the given range of \mbox{50--500 Hz}. In~scenarios with completely unknown sampling rates, the~JITR approach may be the better choice. For~further evaluations, we consider \textit{NN-TA}.

\begin{figure}[H]
\vspace{-9pt}
    \includegraphics[width=0.75\textwidth]{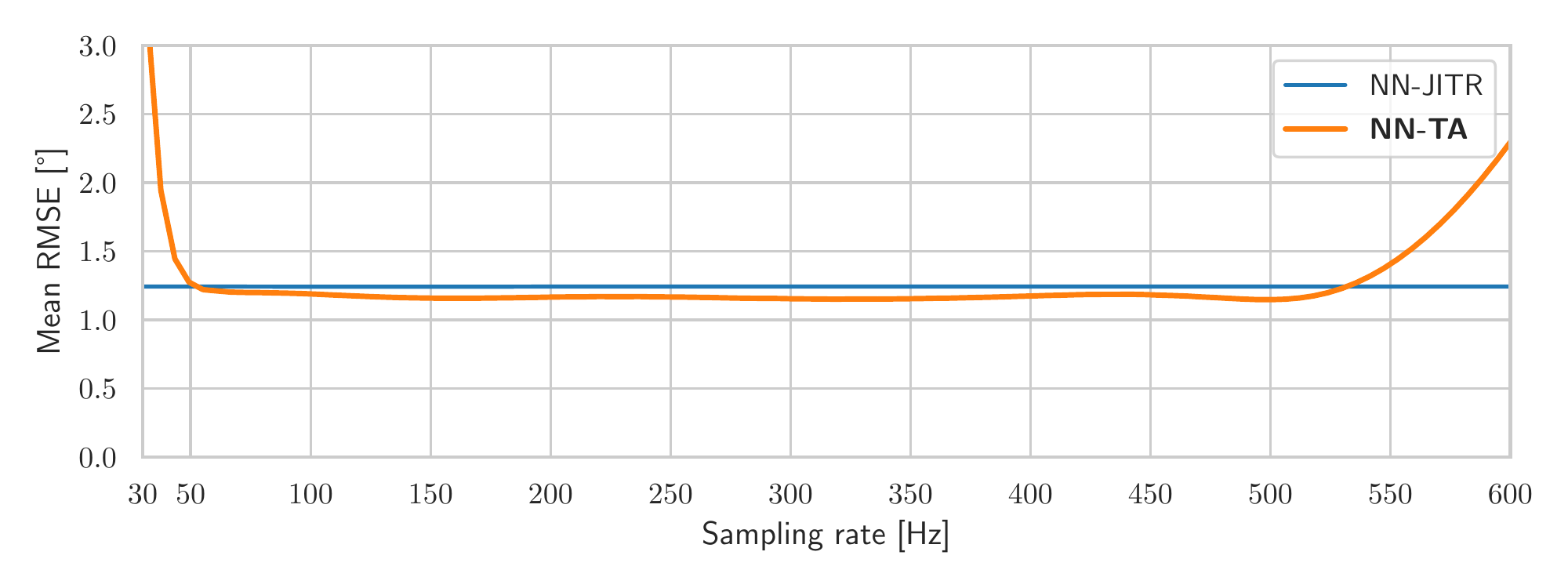}
    \caption{Comparison of the mean RMSE of \textit{NN-JITR} and \textit{NN-TA} over an extended sampling frequency range. \textit{NN-TA} performs slightly better within the trained frequency range but fails outside. Due to just-in-time resampling, \textit{NN-JITR} shows stable performance over the entire extended frequency range. Bold font indicates the chosen final~configuration.}
    \label{fig:gae_2_sampling_comp}
\end{figure}

\subsection{Network Size~Analysis}
We now analyze the effect of the network size on the estimation error and required resources. To~this end, \textit{NN-TA} is trained with a hidden layer size in the range of 10 to 300 on the training dataset and evaluated on the validation~dataset.

Figure~\ref{fig:gae_3_size_parameters} visualizes the influence of the network size on the estimation error. It also shows the exponential relationship between the number of trainable parameters and the neurons per layer. The~estimation error keeps decreasing with the increase of the network size, as~expected. For~the decision of which network size to choose for the final network, we need to consider the trade-off between the increasing computational requirements and gains in estimation~accuracy.

\begin{figure}[H]
\vspace{-6pt}
    \includegraphics[width=0.75\textwidth]{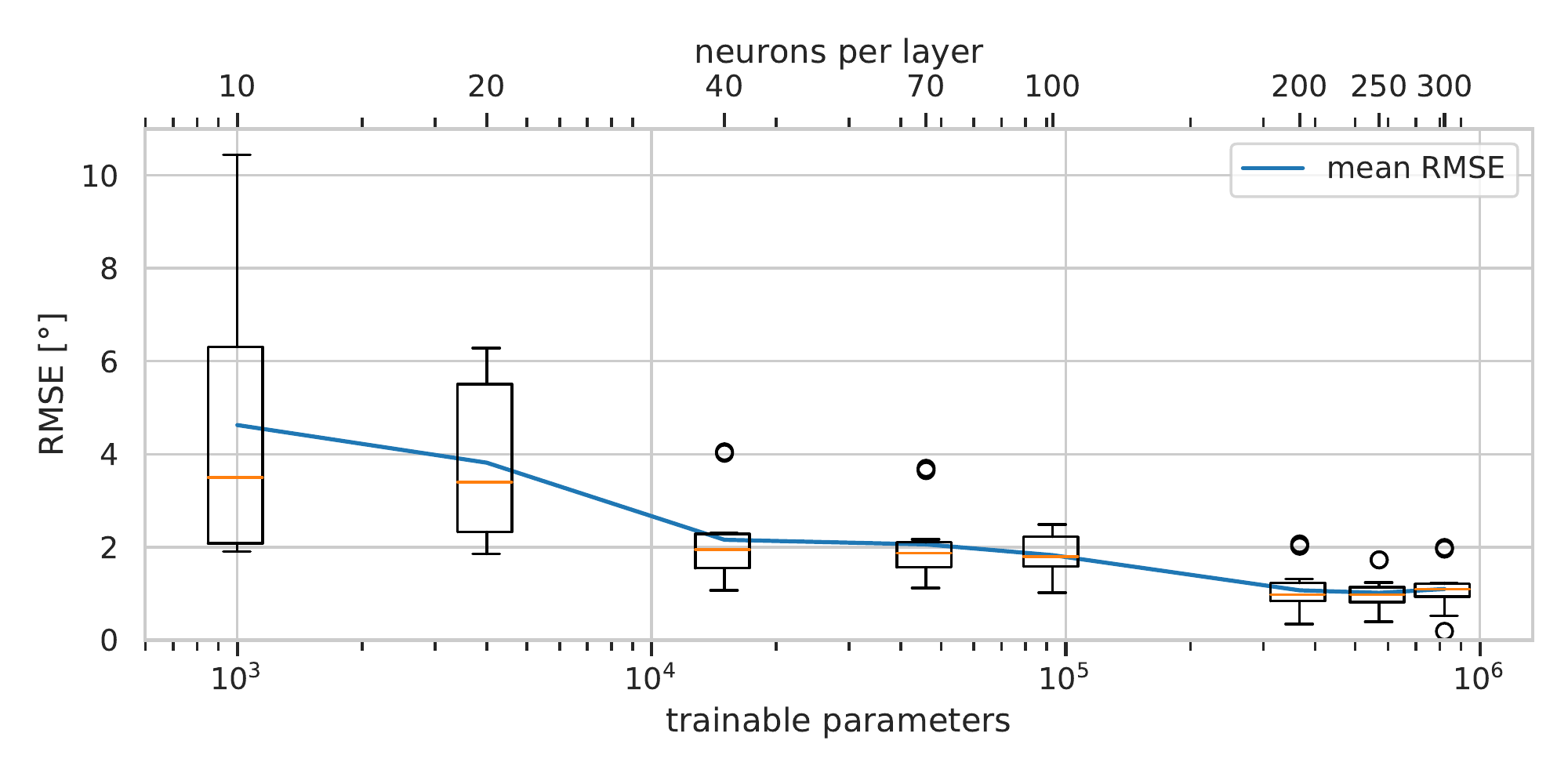}
    \caption{Performance of \textit{NN-TA} for varying network size. Performance is quantified by box plots and the mean of the RMSE over all validation sequences. Increasing the network size leads to continuous performance improvements at the cost of an exponentially growing number of~parameters.}
    \label{fig:gae_3_size_parameters}
\end{figure}

The attitude estimation task often comes with real-time requirements that call for an algorithm that is fast enough to run on the limited hardware of an embedded system. For~analysis of the required resources of \textit{NN-TA} with different sizes, we evaluated the execution on a Jetson Nano~\cite{noauthor_jetson_2019}, which is a microcontroller with an integrated GPU for hardware acceleration of neural networks. On~this platform the models may be executed on the CPU, representing a commonly available, fast microcontroller, or~on the GPU, representing a more expensive embedded system that is specialized for the execution of neural networks. The~prediction times are compared to the ones of a C implementation~\cite{noauthor_open_nodate} and a native Python implementation~\cite{garcia_mayitzinahrs_2021} of two commonly used attitude filters~\cite{mahony_nonlinear_2008}. 

Figure~\ref{fig:gae_3_size_inference} visualizes the results of the study. The~estimation latency depends on the complexity of the estimator as well as on the implementation. While a non-optimized Python implementation of Filter-A is even slower than \textit{NN-TA} with over 800,000 parameters, an~optimized C implementation is orders of magnitude faster. The~choice of the number of parameters is essential for the inference speed of the neural network and for the required memory but has no impact on the ease of use of the final model, which will be applied plug-and-play without any change of parameter values. Considering that the error decreases significantly up to 200 neurons per layer, and~considering the high performance of modern microcontrollers, we chose 200 neurons per layer for the final network, and~we denote this neural network by \textit{RIANN}. With~367,000 fitted parameters, \textit{RIANN} has an estimation latency of 0.29 ms on the target hardware, which results in a high inference speed of 3424 Hz (fast enough for real-time~applications).

\begin{figure}[H]
\vspace{-6pt}
    \includegraphics[width=0.75\textwidth]{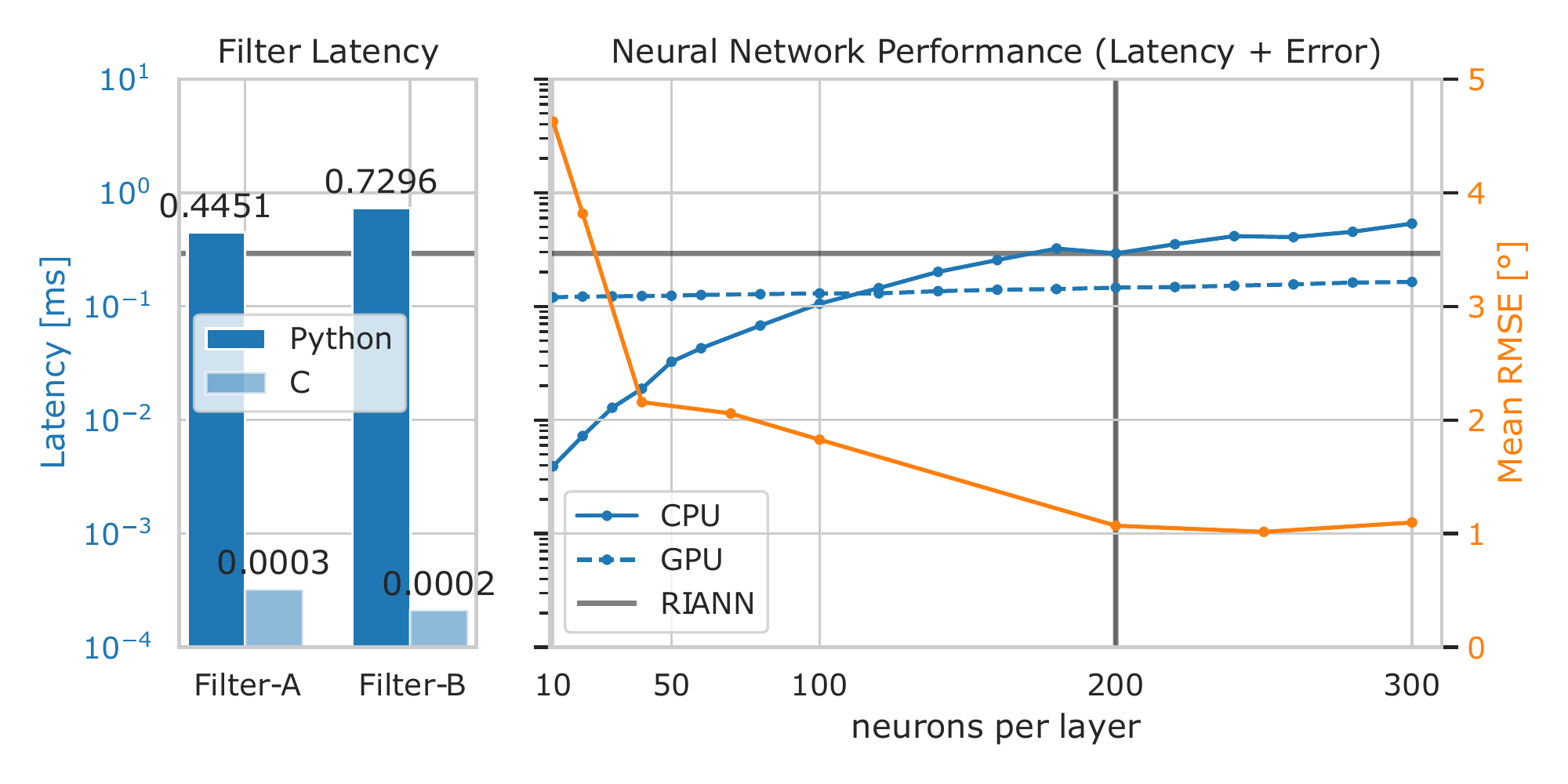}
    \caption{Inference latency and mean RMSE of different neural network sizes, compared to the latency of a C and native Python implementation of popular conventional filters. On~both CPU and GPU, the~network RIANN with 200 neurons per layer is slightly faster than the native \mbox{Python~implementation}.}
    \label{fig:gae_3_size_inference}
\end{figure}

\textit{RIANN} has been exported to the ONNX format to be executed with the ONNX Runtime~\cite{noauthor_onnx_2021}, which is available for a broad range of platforms and hardware. It supports an optimized C implementation for execution on the CPU of the Jetson Nano and a CUDA Version for the GPU. At~low network sizes, the~CPU implementation has smaller latencies than the GPU version because of the CUDA inference overhead. However, with increasing network size, the~GPU latency increases only slightly because the bigger matrix multiplications can be calculated in parallel, for~which the GPU is~optimized. 

\clearpage
\section{Performance~Evaluation}\label{sec:genAnalysis}

The proposed neural-network-based estimator \textit{RIANN} can be considered a viable alternative to conventional attitude estimation filters only if it performs well on data from a broad range of applications with different motion characteristics, environmental conditions, sensor hardware, and~sampling rates. We compare the performance of \textit{RIANN} to the performance of two attitude estimation filters, which are the best performing attitude estimators with a publicly available C implementation according to a recent study of ten different estimators on a dataset with a wide variety of motions~\cite{caruso_analysis_2021}. We denote these filters by Filter-A~\cite{madgwick_ecient_nodate} and Filter-B~\cite{mahony_nonlinear_2008}. While we will, at~some point, optimize the parameters of Filter~A and Filter~B to individual test datasets, we will use \textit{RIANN} always as it is and refrain from performing any additional training or adaptation, to~realistically evaluate its performance on unseen data and unknown application~scenarios.

For the intended comparison, we consider test data from several different datasets as described in Figure~\ref{fig:datasetconfig}, and~we consider three different scenarios that represent different levels of restrictiveness and practical relevance of the assumptions under which the network and the filters are applied to the test~sequences: 
\begin{itemize}
    \item \textit{restrictive scenario:} It is assumed that the sequence starts with a period of perfect rest, during~which the attitude estimation can converge to an accurate estimate before the actual motion starts. Moreover, it is assumed that the turn-on bias of the gyroscopes has been removed in a preprocessing step, which requires a sufficiently long rest phase.
    \item \textit{partially restrictive scenario:} We still assume a rest phase prior to the motion onset, but~no turn-on bias correction has been conducted. We emulate this scenario by adding a random constant bias, which is drawn from a zero-mean normal distribution with a standard deviation of $0.5 ^\circ /s$, to~the bias-free test sequences of the restrictive scenario. 
    \item \textit{realistic scenario:} The sensor already moves when it is turned on and the attitude estimation is started. The~test sequences have the same gyroscope bias as in the partially restrictive scenario, but~the initial rest periods are removed.
\end{itemize}

\textls[-25]{Those scenarios are chosen because these two assumptions, which make the difference between the restrictive and the realistic scenario, are crucial for the practical usability of attitude estimators in many applications, cf. Section~\ref{sec:problem}. In~fact, comparison between these scenarios exposes a common tuning dilemma of conventional filters, as~illustrated in Figure~\ref{fig:gae_4_init_error} for Filter-A. A~low filter gain yields a smaller long-term error, while a high gain yields more rapid initial convergence. This issue can be addressed by initializing the filter with an attitude calculated from the first accelerometer measurement rather than using a fixed initial quaternion. However, without~an initial rest phase, this initialization is inaccurate, and~the same dilemma occurs, cf. Figure~\ref{fig:gae_4_init_error}b. In~summary, despite accelerometer-based initialization, the~low-gain filter needs several seconds up to minutes to converge but then achieves a small error, whereas the filter with a higher gain converges within seconds but exhibits larger errors in the long run. The~same trade-off is observed in other filters, such as Filter-B, and~similar trade-offs and dilemmas are found when balancing between fast and slow or between rotational and translational motions. It is one major goal of this study to investigate whether \textit{RIANN} can overcome these~limitations.}

Figure~\ref{fig:gae_4_box_plots} shows the distribution of the attitude RMSE over all test sequences, grouped by the dataset, in~all three scenarios for \textit{RIANN} and both conventional filters. All test sequences are evaluated with the original dataset-specific sampling frequency in which they were recorded, cf. Figure~\ref{fig:datasetconfig}. The~filters are evaluated in two variants: one with parameters that were numerically optimized on the training data and one with parameters that were optimized for the specific test dataset. The~latter simulates the theoretical best-case in which the circumstances of the specific application are known and ground truth data is available for filter tuning. It grants the filters an advantage that the neural network does not have---RIANN was configured and trained without ever seeing any of the test~data. 

In the \textit{restrictive scenario}, \textit{RIANN} and the conventional filters perform similarly well on most datasets. However, in~the more diverse and dynamic dataset Sassari, the~neural network achieves consistently small errors, while the filter performance is clearly decreased, even for dataset-specific tuning. In~the \textit{partially restrictive scenario} with a realistic gyroscope bias, the~differences become more pronounced. \textit{RIANN} outperforms the conventional filters on at least two of the datasets and consistently maintains mean RMSE values at or below 2 degrees. Finally, in~the \textit{realistic scenario}, \textit{RIANN} clearly outperforms all filter variants in all datasets except EuRoC-MAV, where the errors of all estimators stay similarly small. Especially in datasets that contain highly dynamic motions, the~errors of the conventional filters increase significantly, while the neural network shows no noticeable degradation of~accuracy.

The fact that \textit{RIANN} performs equally well across the different IMU hardware, motion patterns, sampling rates, and~environmental conditions is especially important for all practical applications in which these conditions are unknown or may change over time. In~addition to the improved average performance, it is worth noting that there is not a single sequence with an RMSE of more than $4.5 ^\circ$. This means the worst-case performance of RIANN is clearly better than those of the conventional filters---even if they were tuned for the individual test~dataset.

As a final test, we want to confirm that \textit{RIANN} performs equally well over the whole frequency range. For~this, we resample all test sequences from all datasets to many different frequencies between 50 and 500 Hz and apply \textit{RIANN} to those resampled sequences, while assuming the \textit{realistic scenario}. Figure~\ref{fig:gae_4_nn_fs} visualizes the mean and distribution of the RMSE values over all test sequences plotted over the frequency range. Unsurprisingly, the~performance remains equally good over the entire frequency range. Not only the average but also the maximum errors of the neural network are consistently below the average errors of the conventional~filters.


\end{paracol}
\nointerlineskip
\begin{figure}[H]
\vspace{-9pt}
    \centering
    \widefigure
    \subfigure[~Start with rest]{\includegraphics[width=0.46\textwidth]{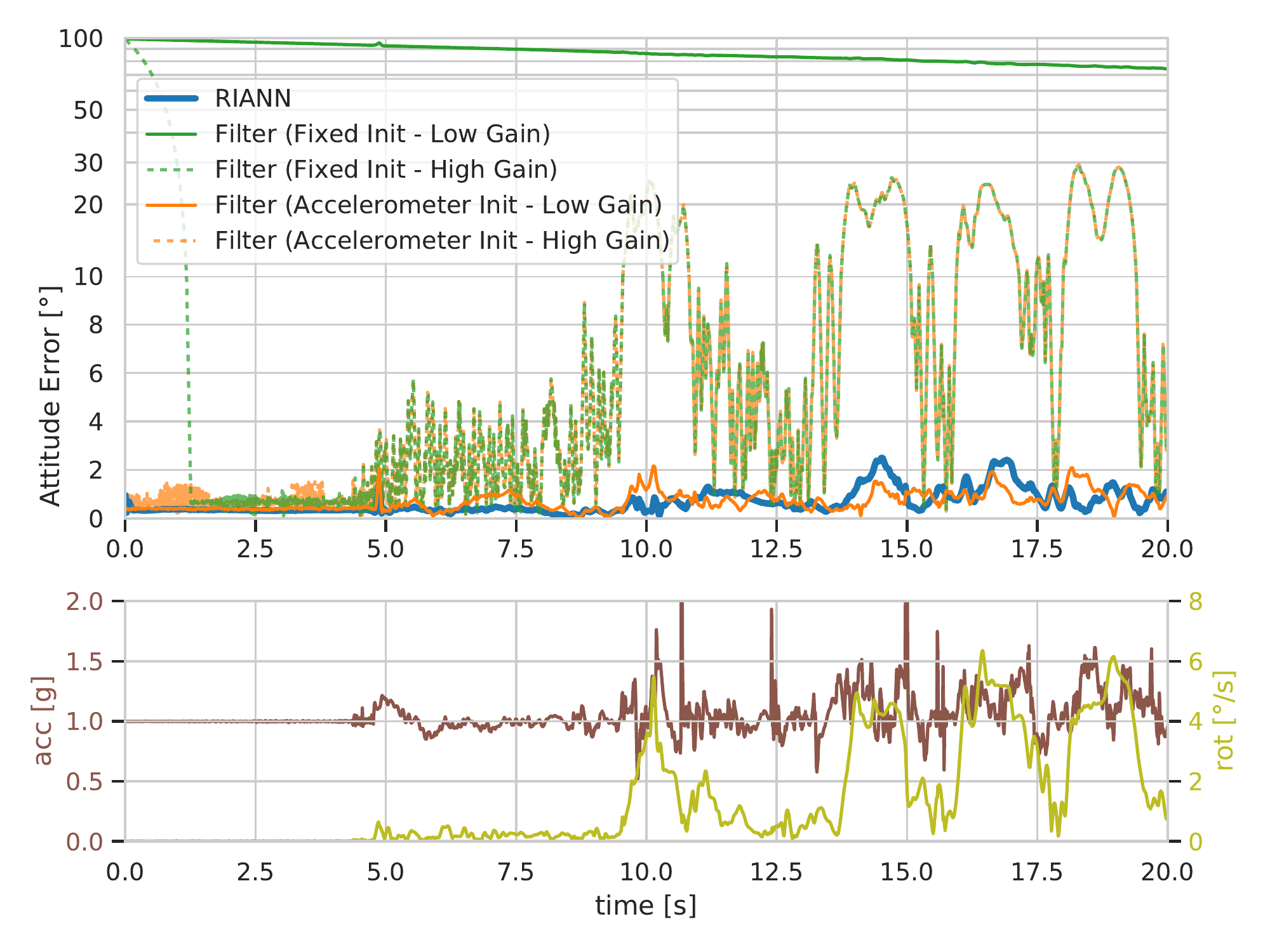}}
    \subfigure[~Start without rest]{\includegraphics[width=0.46\textwidth]{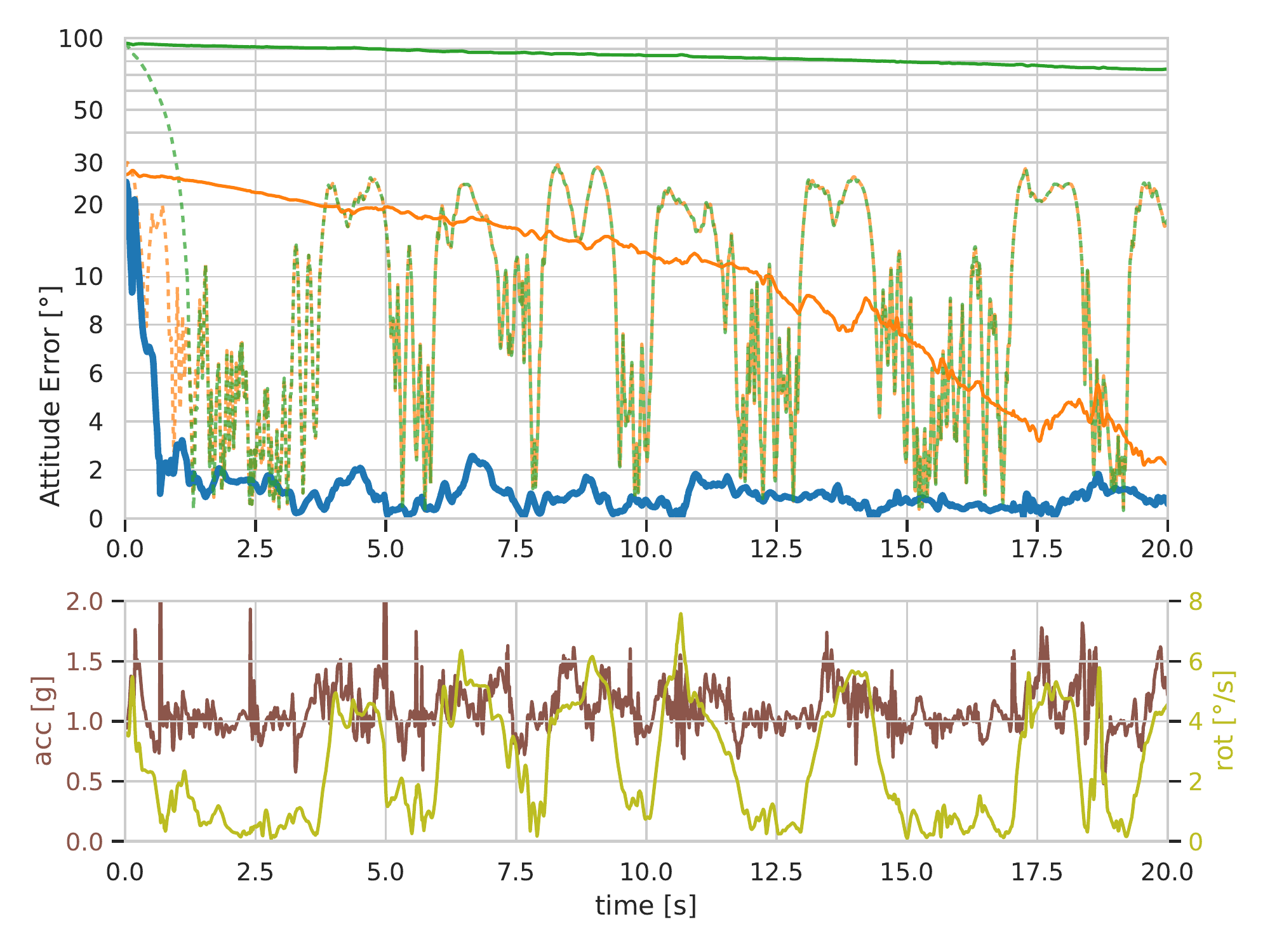}}
    \caption{Attitude error of Filter-A and RIANN in the first 20 s of a sequence starting with rest (\textbf{a}) or without rest (\textbf{b}) with an unknown attitude. Filters exhibit a trade-off between \emph{taking a long time converging to a low error} and \emph{having a larger error over the complete sequence}. In~sequence (\textbf{a}), which starts with rest, only one filter configuration provides good results. In~sequence (\textbf{b}), which starts without rest, there is no good performing filter configuration, whereas \textit{RIANN} performs similarly well as in sequence (\textbf{a}).}
    \label{fig:gae_4_init_error}
\end{figure}
\begin{paracol}{2}
\switchcolumn

\begin{figure}[H]
    \subfigure[~\textit{restrictive scenario} (non-biased + initial rest)]{\includegraphics[width=0.65\textwidth]{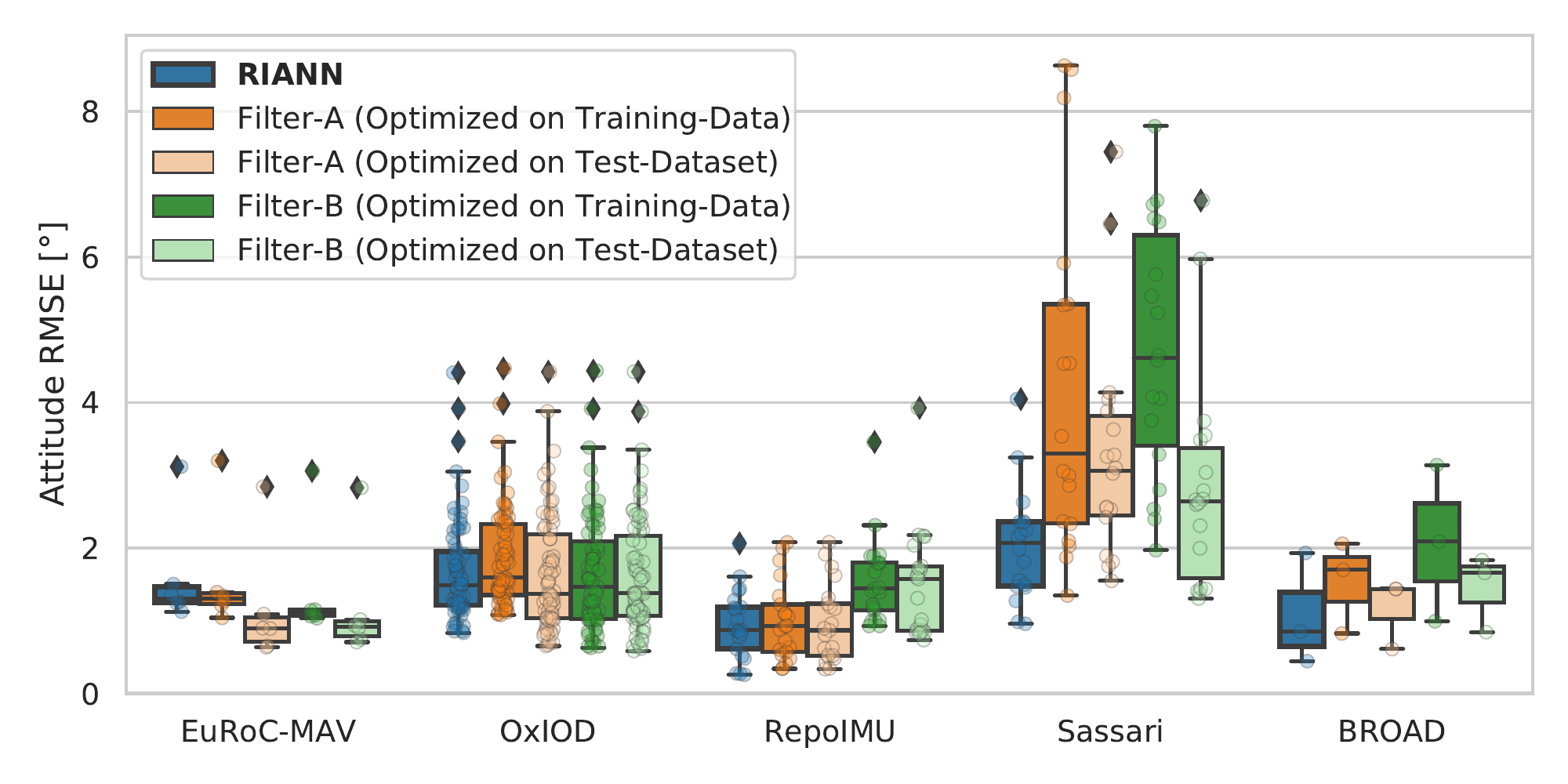}}
    \subfigure[~\textit{partially restrictive scenario} (biased + initial rest)]{\includegraphics[width=0.65\textwidth]{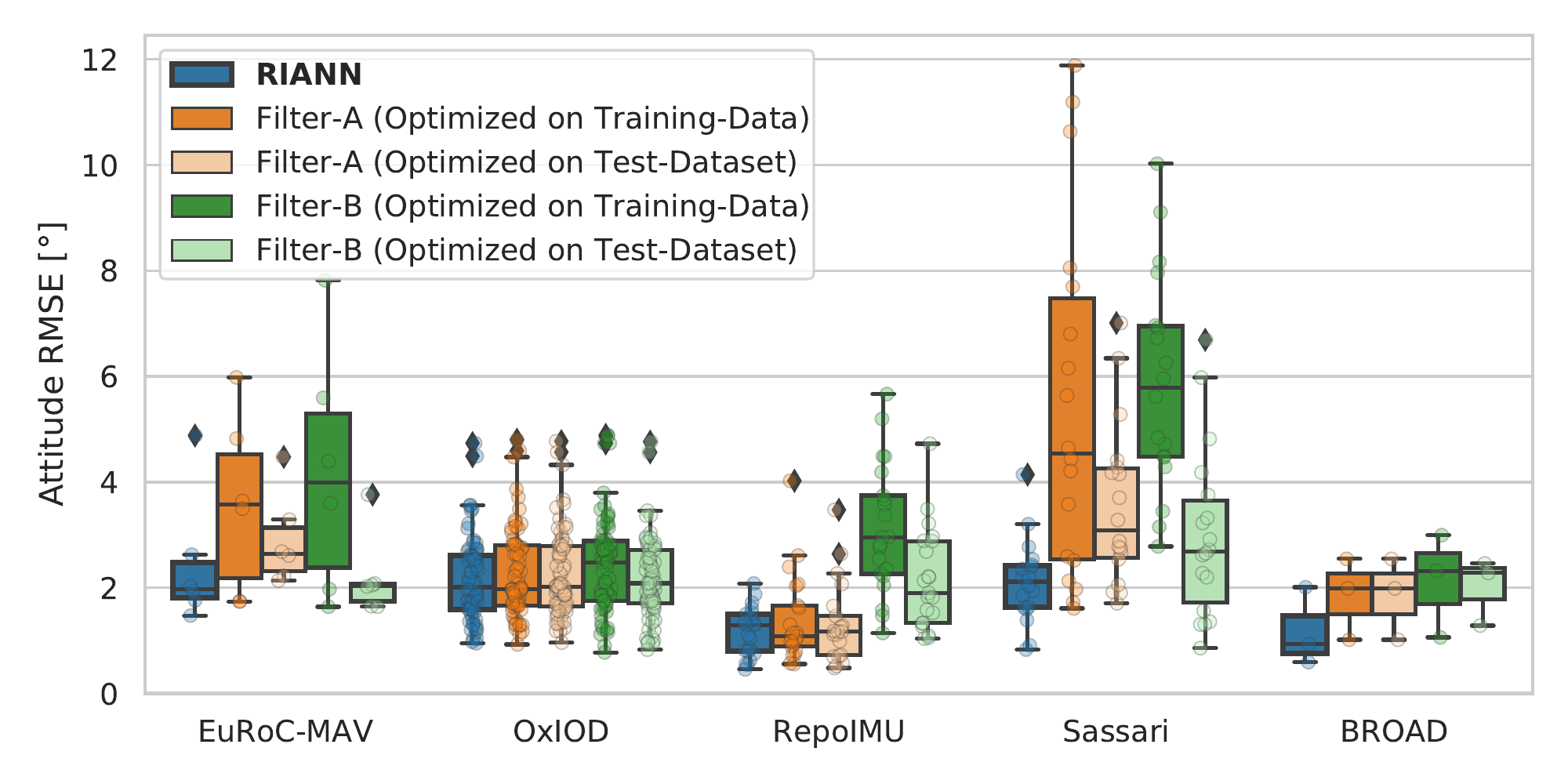}}
    \subfigure[~\textit{realistic scenario} (biased + immediate start)]{\includegraphics[width=0.65\textwidth]{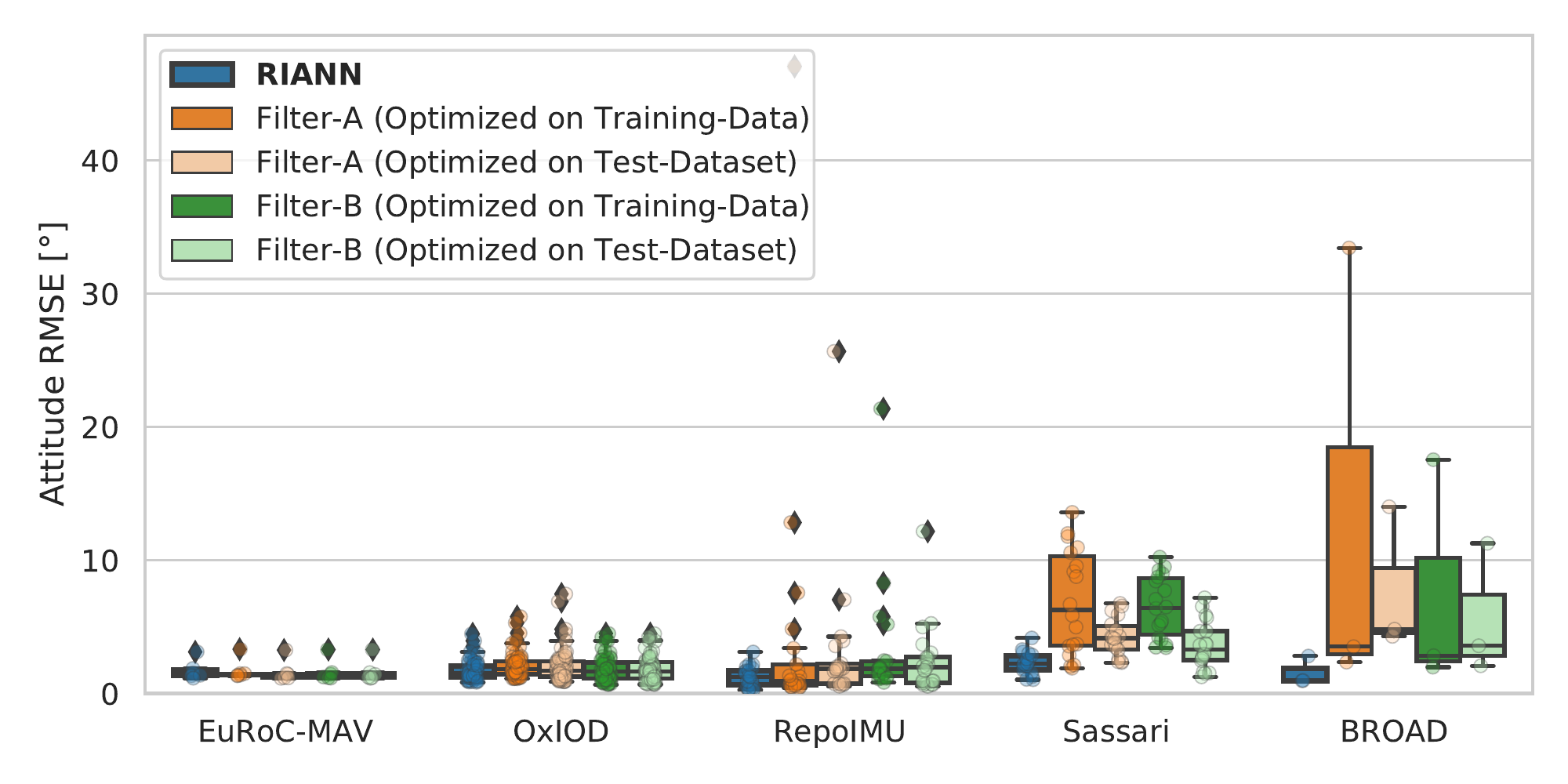}}
    \caption{Comparison of \textit{RIANN} with Filter-A and Filter-B, which are optimized either on the entire training data or on the specific test dataset. Across all scenarios and datasets, \textit{RIANN} performs at least as good as the conventional filters and often outperforms them, especially in the \textit{realistic scenario} and \textit{partially restrictive scenario}.}
    \label{fig:gae_4_box_plots}
\end{figure}

\begin{figure}[H]
    \includegraphics[width=0.75\textwidth]{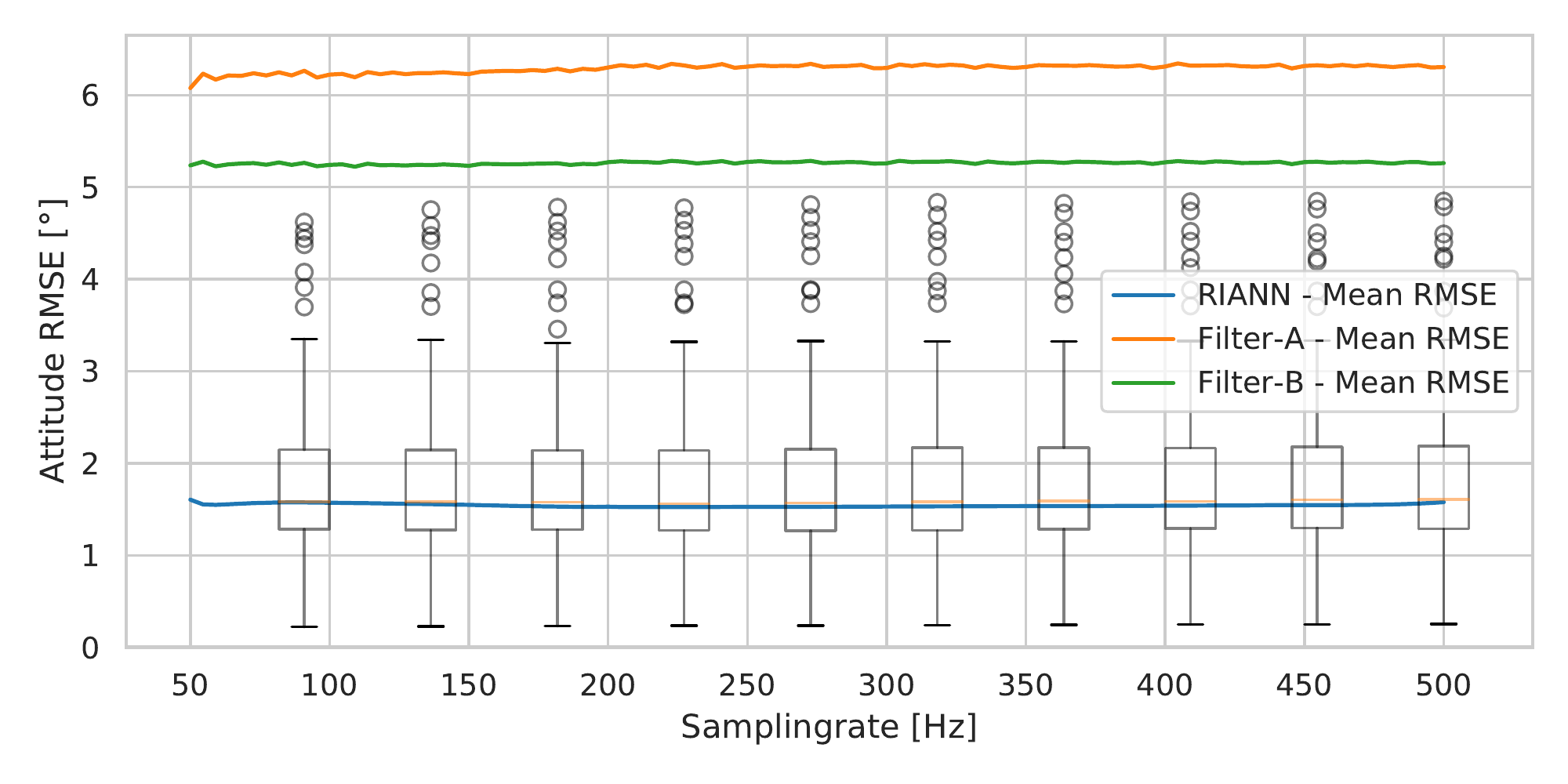}
    \caption{\textit{RIANN's} attitude RMSE distribution (over all test sequences from all datasets) plotted over different sampling rates to which all test data is resampled. The~most challenging, the~\textit{realistic scenario}, is considered. Performance is consistent in the filters and in \textit{RIANN}, with~the latter achieving consistently smaller~errors.}
    \label{fig:gae_4_nn_fs}
\end{figure}

\section{Conclusions}
In this work, we introduced \textit{RIANN}, a~ready-to-use, parameter-free, real-time-capable attitude estimator, which is based on a recurrent neural network with domain-specific advances and trained on two publically available datasets. We compared the performance of \textit{RIANN} with commonly used state-of-the-art attitude filters on a combination of another four publicly available datasets from different~applications.

Our results show that state-of-the-art recurrent neural networks with domain-specific adaptations perform well on the general attitude estimation task over a broad range of specific applications and conditions with no need for retraining or adjustments. \textit{RIANN} even outperforms commonly used state-of-the-art attitude filters in cases, in~which the filter is granted the additional advantage of parameter optimization on the target sequences. Furthermore, \textit{RIANN} has shown a generally low worst-case RMSE of $4.5 ^\circ$ across all test datasets. \textit{RIANN}'s performance generalizes across different hardware, sampling rates, motion characteristics, and~application contexts, which were not included in the training data. This demonstrates that \textit{RIANN} can be expected to perform well in applications with unknown characteristics and conditions and to yield high accuracy without the conventional need for ground truth data recording and context-specific parameter~tuning. 

Compared to conventional filters, \textit{RIANN} requires more computational resources but can still be run in real-time applications on fast, commonly available microcontrollers without specialized hardware. The~proposed domain-specific advances alter the training process but not the neural network implementation itself. This means that \textit{RIANN} can be applied to a wide range of devices using the ONNX format and that platform-specific hardware acceleration capabilities can be exploited. \textit{RIANN} is publicly available at~\cite{weber_riann_2021}.

Future work will be concerned with embedding \textit{RIANN} into motion tracking and analysis toolchains in various applications. Furthermore, the~proposed methods may be extended to the 9D inertial sensor fusion task, which incorporates magnetometer data. Another interesting aspect would be the use of neural architecture search to find the smallest optimized neural network structure that yields competitive performance for applications where the computation capacities are severely limited.
\vspace{6pt} 



\authorcontributions{Conceptualization, methodology, validation, investigation, visualization, and~writing---original draft preparation, D.W. and T.S.;  writing---review and editing, D.W., C.G., and~T.S.; supervision, C.G. and T.S.; software, D.W.; funding acquisition, C.G. All authors have read and agreed to the published version of the~manuscript.}

\funding{This work was partly funded by the German Federal Ministry of Education and Research (FKZ: 16EMO0262). We acknowledge support by the German Research Foundation and the Open Access Publication Fund of TU Berlin.}




\dataavailability{Publicly available datasets were analyzed in this study. This data can be found here: \cite{laidig_broad_nodate, schubert_tum_2018, burri_euroc_2016, caruso_mimu_optical_sassari_dataset_2020, chen_oxiod_2018, szczesna_reference_2016}.}


\conflictsofinterest{The authors declare no conflict of interest. The~funder had no role in the design of the study; in the collection, analyses, or~interpretation of data; in the writing of the manuscript, or~in the decision to publish the~results.} 

\end{paracol}
\reftitle{References}

\end{document}